\documentclass{article}

\usepackage{arxiv}
\usepackage[normalem]{ulem}
\usepackage{subcaption}
\usepackage{pbox}
\usepackage{booktabs}

\usepackage[utf8]{inputenc} % allow utf-8 input
\usepackage[T1]{fontenc}    % use 8-bit T1 fonts
\usepackage{hyperref}       % hyperlinks
\usepackage{url}            % simple URL typesetting
\usepackage{booktabs}       % professional-quality tables
\usepackage{amsfonts}       % blackboard math symbols
\usepackage{nicefrac}       % compact symbols for 1/2, etc.
\usepackage{microtype}      % microtypography
\usepackage{lipsum}
\usepackage{graphicx}
\usepackage[table,xcdraw]{xcolor} % told to include by online table maker
\graphicspath{ {./images/} }

\usepackage{amsmath}

\usepackage[table]{xcolor}
\definecolor{light-gray}{gray}{0.95}
\definecolor{dark-gray}{gray}{0.3}

\usepackage{stmaryrd} % for short arrows
\usepackage{scalerel} % For squishing transition notation into 1 line

\newcommand{\beginsupplement}{%
        \setcounter{table}{0}
        \renewcommand{\thetable}{S\arabic{table}}%
        \setcounter{figure}{0}
        \renewcommand{\thefigure}{S\arabic{figure}}%
     }

\newcommand{\tran}[3] {$\texttt{#1} {\scaleobj{0.7}{\xrightarrow{#2}}} \texttt{#3}$}

\title{A General Model of Conversational Dynamics and an Example Application in Serious Illness Communication}

\author{
 Laurence A. Clarfeld \\
  Department of Computer Science\\
  University of Vermont\\
  Burlington, VT 05405 \\
  \texttt{Laurence.Clarfeld@uvm.edu} \\
  \And
 Donna M. Rizzo \\
  Department of Civil \& Environmental Engineering\\
  University of Vermont\\
  Burlington, VT 05405 \\
  \texttt{Donna.Rizzo@uvm.edu} 
   \And
   Robert Gramling \\
   Department of Family Medicine \\
  University of Vermont\\
  Burlington, VT 05405 \\
  \texttt{robert.gramling@med.uvm.edu} 
     \And
 Margaret J. Eppstein \\
  Department of Computer Science\\
  University of Vermont\\
  Burlington, VT 05405 \\
  \texttt{Maggie.Eppstein@uvm.edu} \\
  \\
}

\begin{document}
\maketitle
\begin{abstract}

Conversation has been a primary means for the exchange of information since ancient times. Understanding patterns of information flow in conversations is a critical step in assessing and improving communication quality. In this paper, we describe \textbf{CO}nversational \textbf{DY}namics \textbf{M}odel (CODYM) analysis, a novel approach for studying patterns of information flow in conversations. CODYMs are Markov Models that capture sequential dependencies in the lengths of speaker turns. The proposed method is automated and scalable, and preserves the privacy of the conversational participants. The primary function of CODYM analysis is to quantify and visualize patterns of information flow, concisely summarized over sequential turns from one or more conversations. Our approach is general and complements existing methods, providing a new tool for use in the analysis of any type of conversation. As an important first application, we demonstrate the model on transcribed conversations between palliative care clinicians and seriously ill patients. These conversations are dynamic and complex, taking place amidst heavy emotions, and include difficult topics such as end-of-life preferences and patient values. We perform a versatile set of CODYM analyses that (a) establish the validity of the model by confirming known patterns of conversational turn-taking and word usage, (b) identify normative patterns of information flow in serious illness conversations, and (c) show how these patterns vary across narrative time and differ under expressions of anger, fear and sadness. Potential  applications of CODYMs range from assessment and training of effective healthcare communication to comparing conversational dynamics across language and culture, with the prospect of identifying universal similarities and unique ``fingerprints'' of information flow. 
\end{abstract}

% keywords can be removed

\section*{Introduction}

Conversation is a fundamental form of human communication. Conversations are highly complex phenomena \cite{prevignano2003discussing}, but use simple rules to maintain discourse \cite{sacks1978simplest}. Humans have an innate ability to learn spoken language from infancy, yet despite the importance of conversations in our daily lives, achieving effective communication through conversations can be difficult \cite{coiera2000conversation, street2009does}.  Developing a better understanding of information flow in different conversational contexts can help guide efforts to improve conversation quality.

Conversation analysis (CA) became established as a discipline of study beginning in the late 1970's with the seminal work of Harvey Sacks and others, such as in their formative 1978 paper \cite{sacks1978simplest}, in which they present a framework for the process of conversation. Sacks described conversation as being highly structured around turn-taking, with participants being able to fluidly transition between turns without turns overlapping. This fundamental property of conversation has subsequently been observed and measured across languages and cultures, revealing itself to be a universal trait \cite{stivers2009universals}. 
Sacks theorized that this discourse is maintained by a set of rules, or norms, that are followed by participants and govern when the speaking floor is relinquished and which participant may speak next. Perhaps the most important aspect of his framework is the heavy focus on the sequential nature of conversation and the dependence of each speaker turn on the turns that came before it \cite{sacks1978simplest}. 

The traditional conversation analytic approach to understanding sequence is to use meticulous transcriptions of recorded conversations in order to study a pre-specified conversational phenomenon and understand its normative patterns  \cite{hutchby2008conversation, hoey2017conversation}. For example, in one early, influential study, Schegloff examined the opening sequence of turns in 500 telephone calls and attempted to explain the patterns he observed \cite{schegloff1968sequencing}. This approach has been widely adopted and applied in diverse contexts, resulting in a vast body of work comprising thousands of research papers. While the value and importance of this inherently qualitative approach remains relevant today \cite{hutchby2008conversation, hoey2017conversation, bolden2015transcribing}, quantitative methods have gained increasing popularity in CA (e.g., \cite{westerman2011conversation, kasper2014conversation}). In particular, Markov Models (MMs)  inherently model sequential events \cite{gagniuc2017markov}, and so have been widely applied in CA. 

In a MM, the likelihood of a given event occurring is determined by the current state of the system, and when an event does occur it causes the system to transition to a new state \cite{gagniuc2017markov}.  The ``order'' of a MM defines the number of previous events that are recorded in each state (i.e., the length of the ``memory'' in the model).  In most CA applications of MMs, state transitions are defined to take place between some constant, fixed intervals of time. Examples of these include 1\textsuperscript{st}-order MMs that were used to classify dialog scenarios in conversations based on speech/silence states \cite{cristani2011generative}, and for identifying conversational structure within non-verbal states, such as gaze patterns in four-person conversations \cite{otsuka2005probabilistic}, and 2\textsuperscript{nd}-order MMs used to study the effects of conversational speech/silence patterns on communication systems \cite{kekre1977two}. In contrast, sequences of speaker order in  four-person conversations were used to predict who the next speaker would be, using MMs up to order 5 \cite{parker1988speaking};  2\textsuperscript{nd}-order MMs proved significantly better than 1\textsuperscript{st}-order MMs for this task, but little further improvement was gained by moving to higher order models on this data set, and a simple context-sensitive model based on speaker roles was shown to out-perform the MMs.

While the examples above use MMs as a tool for making predictions or classifications regarding conversations, another approach to understanding the structure of information flow involves classifying units of conversation by their functional roles (e.g., \cite{thomas1982conversational, core1997coding}).  Once these functional roles have been defined, 1\textsuperscript{st}-order Markov models have been used to understand the sequence of these functions in conversation \cite{thomas1985conversational, thomas1987describing, stolcke2000dialogue}. Influence modeling is yet another Markov-based method, where individual Markov chains for each speaker are coupled together to understand how speakers interact, including understanding which speakers are most influential \cite{basu2001learning, pentland2004social, pentland2004learning} and the functional role of each speaker \cite{dong2007using}. 

Visualizing data to aid interpretation has also grown in popularity \cite{beck2017taxonomy}, and a number of methods have been proposed for visualizing conversational dynamics \cite{wulvik2017temporal, bergstrom2007seeing, angus2011conceptual}. Good visualizations allow complex data to be displayed in a more easily digestible format, allowing readers to better recognize and understand patterns that may not otherwise be apparent \cite{thomas2006visual}. One popular tool for visualizing conversational dynamics is Discursis \cite{angus2011conceptual}, which uses conceptual recurrence plots for unsupervised identification and visualization of shared content between speaker turns in the analysis of conversational discourse. This method has been used in a variety of contexts, including the study of healthcare conversations \cite{fusaroli2016investigating, wong2020advanced, angus2012visualising, angus2013making, watson2015communication, baker2015visualising}. Discursis visualizations portray the lengths of each speaker turn throughout an individual conversation, along with the amount of overlap in content in these turns. Thus, the size and complexity of Discursis visualizations vary for each conversation, and access to full transcriptions are necessary to create these very detailed visualizations of individual conversations.

In this paper, we describe the \textbf{CO}nversational \textbf{DY}namics \textbf{M}odel (CODYM), a novel approach for analyzing and visualizing information flow across sequences of turns in one or many conversations, using 2\textsuperscript{nd}-order and 3\textsuperscript{rd}-order MMs of discretized turn lengths.  CODYMs are based on the assumption that the length of a speaker turn is a simple proxy for the capacity of information conveyed in the turn, and that the amount of information conveyed during a given turn is influenced by the amount of information conveyed in previous speaker turns.  The proposed method is scalable and can be fully automated. Since CODYMs do not rely on knowledge of the specific content of conversations, they do not compromise the privacy of the conversational participants. However, CODYMs can also be contextualized to study information flow patterns surrounding specific topics of interest, if such are known. In contrast to previous applications of MMs to conversation analysis, the primary function of CODYM analysis is to quantitatively summarize and visualize  information flow patterns throughout one or more conversation(s), rather than to make predictions or classifications. Our approach is general and provides a new tool for use in the analysis of any type of conversation. 

In healthcare communication, and especially in serious illness communication, the quality of doctor-patient conversations can have profound tangible impacts \cite{haidet2003building,bernacki2014communication,tulsky2017research}. Promoting high-quality communication in serious illness healthcare is considered a national priority \cite{institute2015dying, ferrell2007national}, and there is an increasing recognition that automated methods for analyzing clinical conversations, such as turn-taking analysis, could provide useful feedback and insights for improving communication between clinicians and patients \cite{tulsky2017research,ryan2019using}. Thus, as an important first application, we apply CODYM analysis to a corpus of 355 transcribed conversations between palliative care clinicians and seriously ill patients, recorded as part of the Palliative Care Communication Research Initiative (PCCRI) \cite{gramling2015design}. These conversations are dynamic, complex phenomena that take place amidst heavy emotions such as anger, fear, and sadness \cite{gramling2020epidemiology}. They include difficult topics such as end-of-life preferences and values, all while patients endure suffering from the symptoms of their illness. We seek to answer the following questions. What are the normative information flow patterns in serious illness conversations? How do those patterns change during the course of a conversation? Do certain words or topics tend to appear more in one information sharing pattern than another? How does the expression of distressing emotion impact information flow? We show that CODYM analysis provides a quantitative approach, with an intuitive interpretation, that helps to answer these questions. 

The remaining sections of this manuscript are organized as follows. We first describe the methods involved in CODYM analysis, and the PCCRI corpus of palliative care conversations. We then present results of applying CODYM analysis to the PCCRI corpus in various ways that demonstrate the model's versatility, followed by a discussion of the significance of our findings and their relation to the current literature. Finally, we close with some general conclusions and ideas for future work. \\

\section*{Methods} \label{sec:methods}

\subsection*{Conversational dynamics model}

Due to the sequential nature of conversations \cite{sacks1978simplest}, we expect that the amount of information conveyed in each turn influences the amount of information conveyed in subsequent turns. Here, we propose using the number of words in a speaker turn as a simple proxy for the capacity of information that the turn can convey.  We define a \textbf{CO}nversational \textbf{DY}namics \textbf{M}odel (CODYM) to be a Markov Model (MM), where events are speaker turns of a given length and states comprise the lengths of some number (defined by the order of the model) of previous turns. CODYMs thus model the sequential patterns in turn lengths. 

\begin{figure}[!h]
\centering
\includegraphics[scale=0.7]{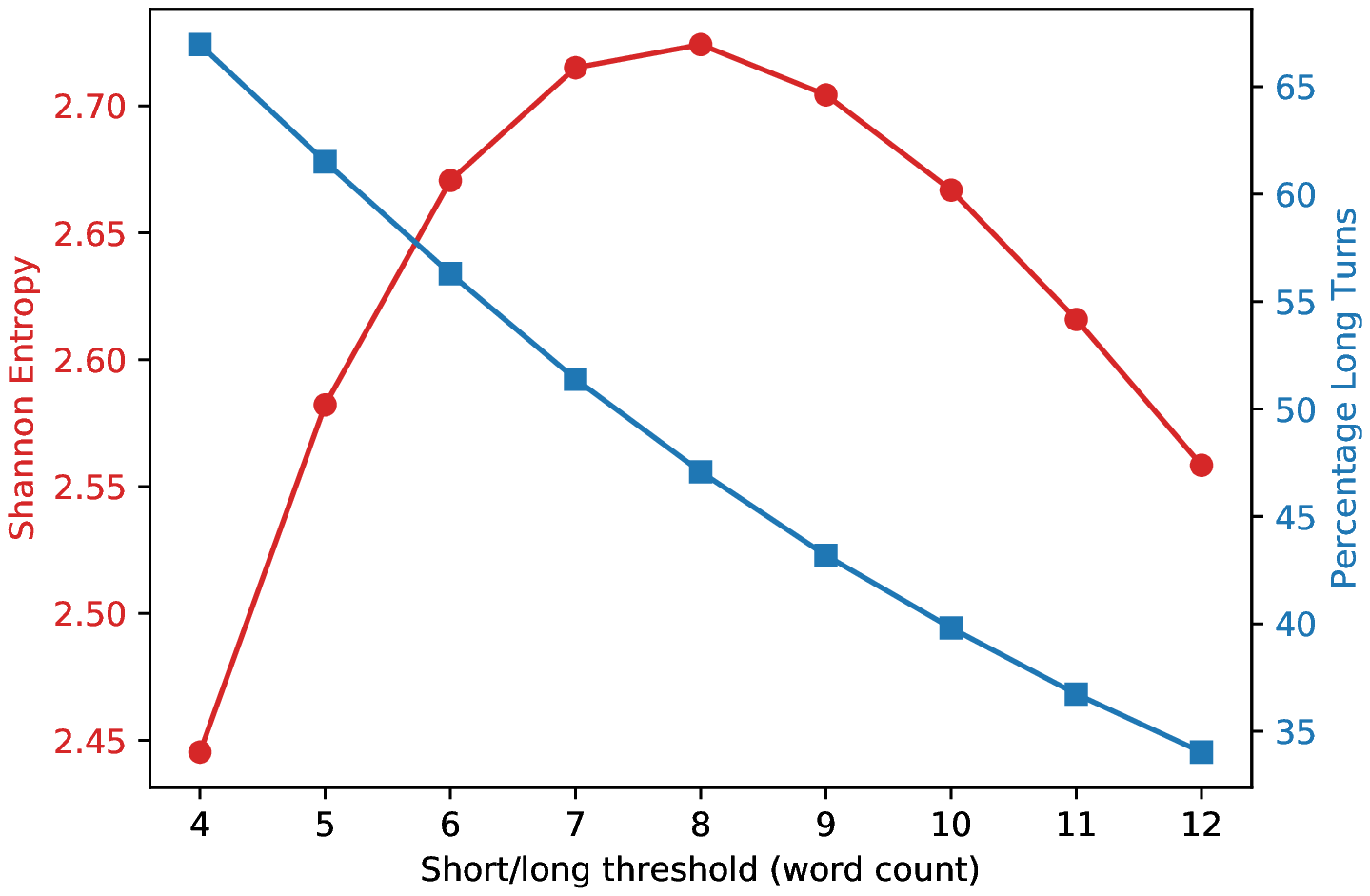}
\caption{\textbf{Binarization of turn length in PCCRI corpus.} Shannon entropy (information content) of transitions (red curve through circles, left $y$-axis) and the percentage of long turns for varying short/long thresholds (blue curve through squares, right $y$-axis) in a 3\textsuperscript{rd}-order CODYM of the PCCRI corpus. Shannon Entropy is calculated $S = \sum_i f_i \log{f_i}$ for the frequency $f_i$ of each transition. The short/long threshold is defined such that for a threshold, $t$, any turn with $t$ or more words is considered long. For all experiments in this study, we define short turns to be 7 or fewer words and long turns to be 8 or more words.}
\label{fig:optimizing_rate}
\end{figure}

Any MM requires a discrete state-space, so turn lengths in a CODYM are discretized into a finite number of bins. Although, in principle, turn lengths can be discretized into any number of bins, the most appropriate number of bins  will depend, in part, on the size of the data set.  In the PCCRI corpus used here, turn lengths follow a heavy-tailed distribution, with a median turn length of 7 (Fig. \ref{fig:words_per_turn}).  Here, we binarize turn lengths into short (\texttt{S}) turns and long (\texttt{L}) turns, with short turns defined as those with 1-7 words and long turns as those with 8 or more words. Using the median turn length as the maximum length of short turns (a) creates a relatively balanced data set with 53,751 short turns and 47,812 long turns, and (b) maximizes the Shannon entropy (a measure of information content) for the distribution of states in a 3\textsuperscript{rd}-order CODYM (Fig. \ref{fig:optimizing_rate}).  In preliminary experimentation with the PCCRI corpus, we found that discretizing turn lengths into ternary bins (Short/Medium/Long) was problematic because it both (a) created sample size issues by reducing the number of turns associated with each transition, and (b) resulted in more complex models that were difficult to interpret.

\begin{figure}[!h]
\centering
\includegraphics[scale=0.44]{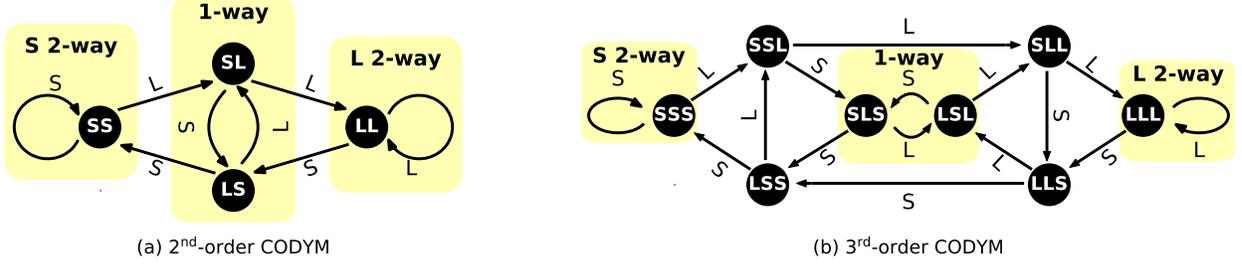}
\caption{\textbf{Network representations of CODYMs.} Network depictions of (a) 2\textsuperscript{nd}-order  and (b) 3\textsuperscript{rd}-order CODYMs, where turn lengths are binarized as short (\texttt{S}) or long (\texttt{L}). Nodes (black circles) represent states that are defined by the lengths of the 2 or 3 previous turns, respectively; edges (arrows) represent transitions between states and are labeled with the length of the turn on that transition. The areas highlighted in yellow represent important sub-networks we refer to as short two-way information exchanges (labeled ``\texttt{S} 2-way''), one-way information exchanges (labeled ``1-way''), and long two-way information exchanges (labeled ``\texttt{L} 2-way'').}
\label{fig:state_space_diagram}
\end{figure}

Note that the number of states in an $N$\textsuperscript{th}-order CODYM of binarized turn lengths has $2^N$ states and $2^{N+1}$ transitions. For example, a 2\textsuperscript{nd}-order CODYM has 4 states and 8 transitions, and a 3\textsuperscript{rd}-order CODYM has 8 states and 16 transitions, as illustrated in Fig. \ref{fig:state_space_diagram}, where states are represented as nodes and transitions are represented as directed edges in a network.  We use 2\textsuperscript{nd}- and 3\textsuperscript{rd}-order CODYMs in our analyses for the reasons explained below.

We use 3\textsuperscript{rd}-order CODYMs for analyzing normative patterns in serious illness conversations, and for examining how these patterns change both temporally and  when distressing emotions are expressed. We selected 3\textsuperscript{rd}-order CODYMs for this because: (a) for dyadic conversations, such as between a patient and a clinician, having memory of 3 previous turns can represent a complete back-and-forth exchange between the 2 sides; (b) higher-order models become difficult to interpret; (c) large state spaces result in fewer observations of each state/transition, potentially resulting in small sample sizes that may preclude accurate characterization of normative patterns (e.g., the median numbers of occurrences of the different states for each conversation in the PCCRI corpus are 57, 28, 13, 6, and 3, for CODYMs of orders 1-5, respectively); and (d) we found that features taken from 3\textsuperscript{rd}-order CODYMs had greater predictive power over those from 1\textsuperscript{st}- or 2\textsuperscript{nd}-order CODYMs on an emotion-based classification task (described in more detail later), as shown in Table \ref{table:class_order}. On the other hand, a 2\textsuperscript{nd}-order CODYM has half the number of transitions as a 3\textsuperscript{rd}-order model (Fig. \ref{fig:state_space_diagram}). Thus, when studying patterns of word associations with transitions, we use 2\textsuperscript{nd}-order CODYMs to increase the number of words associated with each of the transitions. 

Different sub-networks of a CODYM can be interpreted as distinct regimes of information flow (highlighted in yellow  in Fig. \ref{fig:state_space_diagram}).  For example, we refer to the center loop (\tran{\tran{SL}{S}{LS}}{L}{SL}, in a 2\textsuperscript{nd}-order CODYM, and \tran{\tran{SLS}{L}{LSL}}{S}{SLS}, in a 3\textsuperscript{rd}-order CODYM) as ``one-way information exchanges'', because alternation between \texttt{S} and \texttt{L} turns in a dyadic conversation implies that one party is conveying most of the information.  We refer to the leftmost self-loop (\tran{SS}{S}{SS}, in a 2\textsuperscript{nd}-order CODYM, and \tran{SSS}{S}{SSS}, in a 3\textsuperscript{rd}-order CODYM) as ``short two-way information exchanges''. Conversely, we refer to the rightmost self-loop (\tran{LL}{L}{LL}, in a 2\textsuperscript{nd}-order CODYM, and \tran{LLL}{L}{LLL}, in a 3\textsuperscript{rd}-order CODYM)  as ``long two-way information exchanges''.  As we will show, each of these information flow regimes can be associated with different functions in conversations. 

\subsection*{Observing patterns of information flow}

In most MMs, the weights on all outgoing edges of each given node  sum to 1.0, where each edge weight represents the probability that the node is left {\em via} that edge. This is appropriate when MMs are used as generative models or to make predictions of future states. However, with CODYMs our primary intent is to study patterns of information flow through all states and transitions in an existing corpus of dialog, not to generate simulated sequences of short and long speaker turns  or to predict subsequent turn lengths. Thus, in a CODYM, the weights on nodes and edges represent their respective percentage frequencies of occurrence over all turns being analyzed. Consequently, the sum of all edge weights (transition frequencies) in an entire CODYM is 100\% and the sum of all node weights (state frequencies) is also 100\%.
 
We ``populate'' a CODYM by computing observed frequencies of each state/transition across a specified set of speaker turns. This set may comprise all turns in an entire corpus of conversations, all turns for a given speaker, all turns within individual conversations, or some other subset of turns that satisfy some pre-specified condition, depending on the question being addressed. When different CODYMs are populated separately for a number of conversations, they can be visualized as a single CODYM populated with the mean weights for each of states and transitions, averaged over all of the CODYMs populated from individual conversations. A CODYM of mean frequencies can be interpreted as a representation of the overall ``normative'' pattern of information flow in the set of conversations under study, assuming the distributions of frequencies are uni-modal for each state and transition. Alternatively, when CODYMs are populated from a subset of turns in the corpus, turns from multiple conversations may be pooled to increase sample sizes in the states/transitions; in this case, we compute a single set of CODYM frequencies for the pooled data.

Where possible, we seek to determine whether a CODYM computed from observed data exhibits state and/or transition frequency distributions that differ from what would be expected if the given conversational feature of interest was independent of prior sequences of turn lengths (the null hypothesis). How we determine ``expected'' frequencies, and how we compare observed frequencies to them,  depends on the particular experiment. For example, we compare two sets of observed CODYM frequency distributions, computed from turns with or without a given feature present, using 2-sample  Kolmogorov-Smirnov tests. However, in most cases, we generate expected frequencies from appropriate null CODYMs, which are populated from data that have been randomly sampled (or randomly reorganized) in such a way as to disrupt any possible association between the conversational feature of interest and previous sequences of turn lengths, while preserving other salient characteristics of the data, such as sample size and the number of long and short turns used by patients and clinicians.  When comparing to null CODYMs created by random sampling, we derive empirical probability distributions for null models computed from randomly sampled data by generating 1000 random copies of each, using Monte Carlo (MC) simulations. (Prior experimentation had shown that probability distributions were quite stable when created with 1000 MC simulations). When we visualize these null CODYMs, we display mean state and transition frequencies, averaged over all 1000 copies. If observed state and/or transition frequencies are outside of the empirically derived 95\% confidence interval from the distribution of 1000 corresponding null models, the difference is considered to be statistically significant at the $p<0.05$ level. To minimize confusion, we provide the specifics of how we determined expected values for each type of experiment in the Results.

State space diagrams of CODYMs are displayed in one of two formats, based on what we believe most effectively conveys the results we are trying to highlight. In some cases, we display transition frequencies directly ($\%Observed$ or $\%Expected$), while in other cases we display  mean observed transition frequencies minus the mean expected frequencies of the corresponding null models ($\Delta frequency = \%Observed - \%Expected$).  In the former, all state weights sum to 100\%, all transition weights sum to 100\%, and the color bars represent only positive percentages. In the latter, transition weights sum to 0\%, negative values indicate the degree to which the feature of interest is under-represented on a given transition, and positive values indicate the degree to which it is over-represented, relative to the corresponding null model. We use different colormaps to help distinguish these two visualization techniques. In both visualization formats, the thickness and color of transition arrows indicate the magnitude of the corresponding transition weights, and nodes are sized according to the frequency of their respective states. Where relevant, state and transition frequencies that are statistically significantly different from the corresponding null model are indicated by coloring states black ({\em vs.} gray) and drawing transition arrows with solid lines ({\em vs.} dashed lines). 

To assess how normative patterns in information flow may change over the course of a conversation, we divide the turns of each conversation in the corpus into sequential deciles of words (ten bins of narrative time, as in \cite{ross2020story}), stratified by patient and clinician turns. Note that different conversations have different numbers of turns, so the number of turns per bin varies by conversation. Individually, conversations average only about 28 turns per decile, split approximately equally between patient turns and clinician turns. Since 14 turns are inadequate to robustly determine frequencies on 16 transitions, we pool the data by summing the number of patient and clinician turns per decile over all conversations.  We then compute the frequencies of the pooled turns on each of the 16 transitions in 3\textsuperscript{rd}-order CODYMs, one per decile.

\subsection*{Contextualization of CODYMs}\label{sec:MethodsContextualization}
A CODYM is based exclusively on turn lengths, so is thus independent of what is actually said or expressed during those turns. However, a CODYM can be used to examine the information flow patterns involving different words, topics, expressions of emotion, or more generally, ``contextual events''. 

\begin{figure}[!h]
\centering
\includegraphics[scale=0.4]{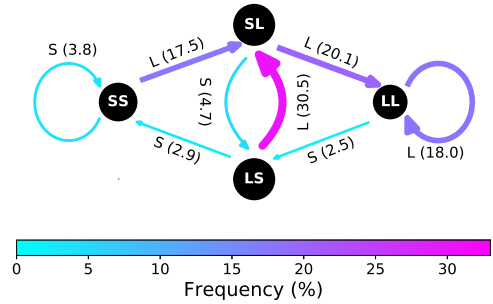}
\caption{\textbf{CODYM of word usage in PCCRI corpus.} Frequencies of occurrence of all words in the PCCRI corpus on transitions of a 2\textsuperscript{nd}-order CODYM. Transition labels indicate the length of the turn on that transition, parenthetically followed by the percentage of word occurrence on that transition. Edge thickness and color indicate $\%Observed$ for each transition, and node diameter indicates $\%Observed$ for each state.}
\label{fig:clusternull}
\end{figure}

We use two complementary approaches to study how word usage varies between conversational regimes. In the first approach, we use an unsupervised cluster analysis to determine which commonly used words have similar CODYM patterns of information sharing. Of the 14,848 unique words that appear in the PCCRI corpus, we only consider those that appear at least 100 times and for which the absolute value of $\%Observed - \%Expected$ is greater than 10; expected values are the observed frequencies in a 2\textsuperscript{nd}-order CODYM populated from all words over all turns in the corpus (Fig. \ref{fig:clusternull}). We use this method for prefiltering words because: (a) requiring a minimum number of occurrences ensures that a word is frequent enough in the corpus to asses its typical usage; and (b) by considering only words whose information flow patterns differ substantially from overall word frequencies, we focus on words that have specialized usages with respect to the normative information flow pattern. 

The resulting list of 114 words was clustered using a standard K-Means clustering algorithm (implemented with Python's scikit-learn package, with default settings \cite{scikit-learn}), using the 8 transition frequencies from  2\textsuperscript{nd}-order CODYMs (computed for each of the 114 words) as the input features for clustering. Note that each of the 114 words are weighted equally during this clustering, even though some words occurred much more frequently than others. After qualitatively assessing cluster membership when using between three and ten clusters, we determined that using six clusters resulted in the most logically cohesive word groupings.

In the second approach, we populate CODYMs using frequencies of occurrence of terms from pre-defined lists. In this work, we focus on two such lists of interest in serious illness conversations: hedging terms and treatment terms. 

There are many definitions of uncertainty terms and subtypes used in clinical conversations (e.g., \cite{vincze2015uncertainty,strekalova2017language,han2011varieties, hanauer2012hedging}). Here we focus on a list of ``hedging'' terms, recently developed for use in natural language processing of serious illness conversations, which was found to be the most prevalent subtype of uncertainty expressed in the PCCRI corpus \cite{durieux_2019}. According to this definition,  hedging occurs in a conversation when the speaker makes a word choice to suggest something rather than state it as fact, such as to say ``I expect you will feel better'' instead of ``You will feel better''.   Using the word \textit{expect} injects some uncertainty into the statement.  Some examples of hedging terms include expect, hint, imply, perceive, presume, speculate, suspect, and think. The complete list of 35 hedging terms used in this study, and the frequency of occurrence of each term in the PCCRI corpus, is shown in Table \ref{tab:hedgingterms}. See \cite{durieux_2019} for a detailed description of how this list was compiled. 

Treatment terms are those used to discuss medical care provided to patients. A list of 51 treatment words occurring at least 100 times each in the PCCRI corpus was previously prepared, and it was shown that these words vary  temporally throughout the PCCRI conversations, being most prevalent in deciles 4 and 5  \cite{ross2020story}. Some examples of treatment terms include antibiotics, ICU, chemo, fentanyl, machine, milligram, procedure, radiation, and ventilator. The complete list of the treatment terms used in this study (from \cite{ross2020story}), and the frequency of occurrence of each term in the PCCRI corpus, are shown in Table \ref{tab:treatmentterms}. 

Note that specific hedging terms or treatment terms that are more frequent in the overall corpus have a greater influence on resulting state and transition frequencies, since CODYM frequencies were computed over all occurrences of all terms in a given list. Thus, to assess robustness of the observed information patterns for both hedging and treatment terms, we remove a random 10\% of terms that occurred at least once in the PCCRI corpus from each list and re-perform the analysis. This process is repeated with a different 10\% of terms removed each time, until all terms have been removed at least two times.

In what is perhaps our most intriguing application of contextualization, we study how the expression of  distressing emotions by patients affects patterns of information flow, both locally (by comparing CODYM patterns between turns in which anger, fear or sadness were or weren't expressed) and globally (by comparing CODYM patterns between entire conversations in which anger and/or fear were or weren't expressed). As part of the latter analysis, a Random Forest classifier (implemented using Python's Scikit-learn package with 100 trees  and default settings \cite{scikit-learn}) was trained to distinguish conversations that included anger and/or fear from those that didn't, using normalized conversation-level transition frequencies from 1\textsuperscript{st}-order through 5\textsuperscript{th}-order CODYMs as input features. Each Random Forest was trained on a random 80\% of conversations from each class and tested on the remaining 20\% of conversations from each class. This process was repeated 1,000 times for each order CODYM, with different random 80-20 data splits. The distribution of observed testing accuracies in these 1000 Random Forests was used to assess how consistently CODYM features could distinguish between conversations with fear and/or anger and those without.

\subsection*{The PCCRI corpus} \label{sec:corpus}

The Palliative Care Communication Research Initiative (PCCRI) is a multisite observational cohort study conducted between January 2014 and May 2016 \cite{gramling2015design}. The study took place at two large U.S. academic medical centers, one in the Northeast and one in the West. Any English-speaking patients who were hospitalized and referred for inpatient palliative care consultation were eligible for this study, provided they were diagnosed with a metastatic nonhematologic cancer, did not have a documented exclusively comfort-oriented plan of care at the time of referral, were age 21 or over, and were able to consent for research either directly or via health care proxy (if lacking capacity as determined by the clinical team). All members of the interprofessional Palliative Care Inpatient Consult teams at both sites were eligible to participate. 
 
A total of 240 hospitalized patients with advanced cancer at the time of referral for inpatient palliative care consultation were enrolled in the study. Four withdrew, three died, and two were discharged before completing the palliative care consultation.   Each consultation comprised one to three conversations between the patient, and potentially family members and/or close friends of the patient, and the palliative care team. More than one conversation occurred with the same patient when the initial conversation was only a preliminary assessment or when a conversation was interrupted prematurely (e.g., a patient was taken for x-rays) \cite{gramling2015design}. 

All conversations that were part of a palliative care consultation were audio recorded. With prior informed consent from all study participants, digital recorders were placed in unobtrusive locations in the rooms where the conversations took place (e.g., on a tray table next to a patient's bed); research assistants retrieved the recorders at the end of the visit by the palliative care team. All audio recordings were later transcribed verbatim and prepared in a standard format to facilitate natural language processing. The speaker during each transcribed turn was tagged as either a patient (which could include family members and/or close friends who were present in support of the patient) or a clinician, except in rare occasions ($< 1\%$ of turns) when transcribers could not determine whether the speaker was from the patient side or the clinician side (these turns were excluded from analyses that stratified turns by patient and clinician).  In total, we examined 360 conversations that were recorded and transcribed for 231 unique patients.

\begin{figure}[!h]
\centering
\includegraphics[scale=0.7]{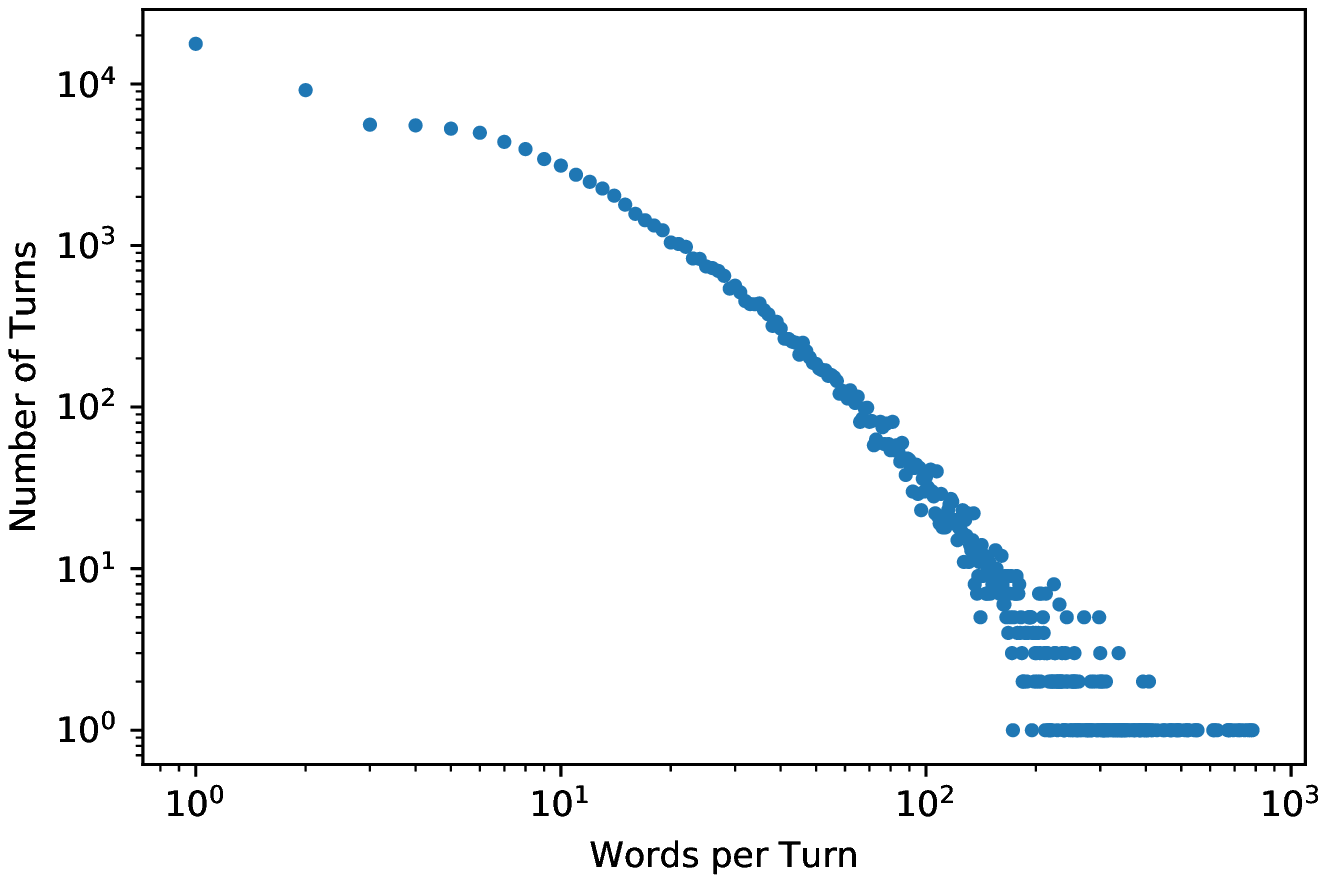}
\caption{\textbf{Words per turn in PCCRI corpus.} The number of words per turn, for each of the 101,563 turns in the PCCRI corpus.}
\label{fig:words_per_turn}
\end{figure}

Of these, five transcripts were excluded from this study because either a high proportion of speaker turns were inaudible rendering the transcripts very incomplete, or because the conversations were too short (less than 20 speaker turns long) to perform meaningful analysis. The remaining 355 conversations, which were used in our analyses, contained 1,464,167 total words (14,812 unique) in 101,563 total speaker turns, with a median of 242 turns per conversation (Fig. \ref{fig:words_per_turn}).

Since transcribers were not always able to distinguish patients from family members or friends, or which clinician was speaking, in this work we adopt the convention that ``patient'' refers to anyone on the patient-side of the conversation and ``clinician'' refers to anyone on the clinical team. Although there were up to 8 participants in a conversation, patient turns were followed by clinician turns (and {\em vice versa}) during 86\% of all transitions, closely resembling the alternating speaker pattern that would be expected in dyads. 

All speaker turns in which the patient was audibly perceived to be expressing anger, fear, or sadness, had been previously labeled in the PCCRI transcripts, using well-established and reliable human coding methods \cite{ingersoll2019contagion, alexander2014emotional}. Anger was defined to include expressions of either frustration or anger.  Fear was defined to be inclusive of words and sounds indicating worry, anxiety, fear or terror. Sadness was defined to include expressions with sad, disappointed, depressed, hopeless or discouraged sentiments. Turns that included multiple sentiments were coded as such. Ambiguous words or sounds that {\em might} indicate underlying emotion, or referred to emotions felt in the past, were not included in our analysis.  

\section*{Results} \label{sec:results}

\subsection*{Normative patterns of information flow in serious illness conversations}\label{sec:normative}

\begin{figure}[!h]
\centering
\includegraphics[scale=0.7]{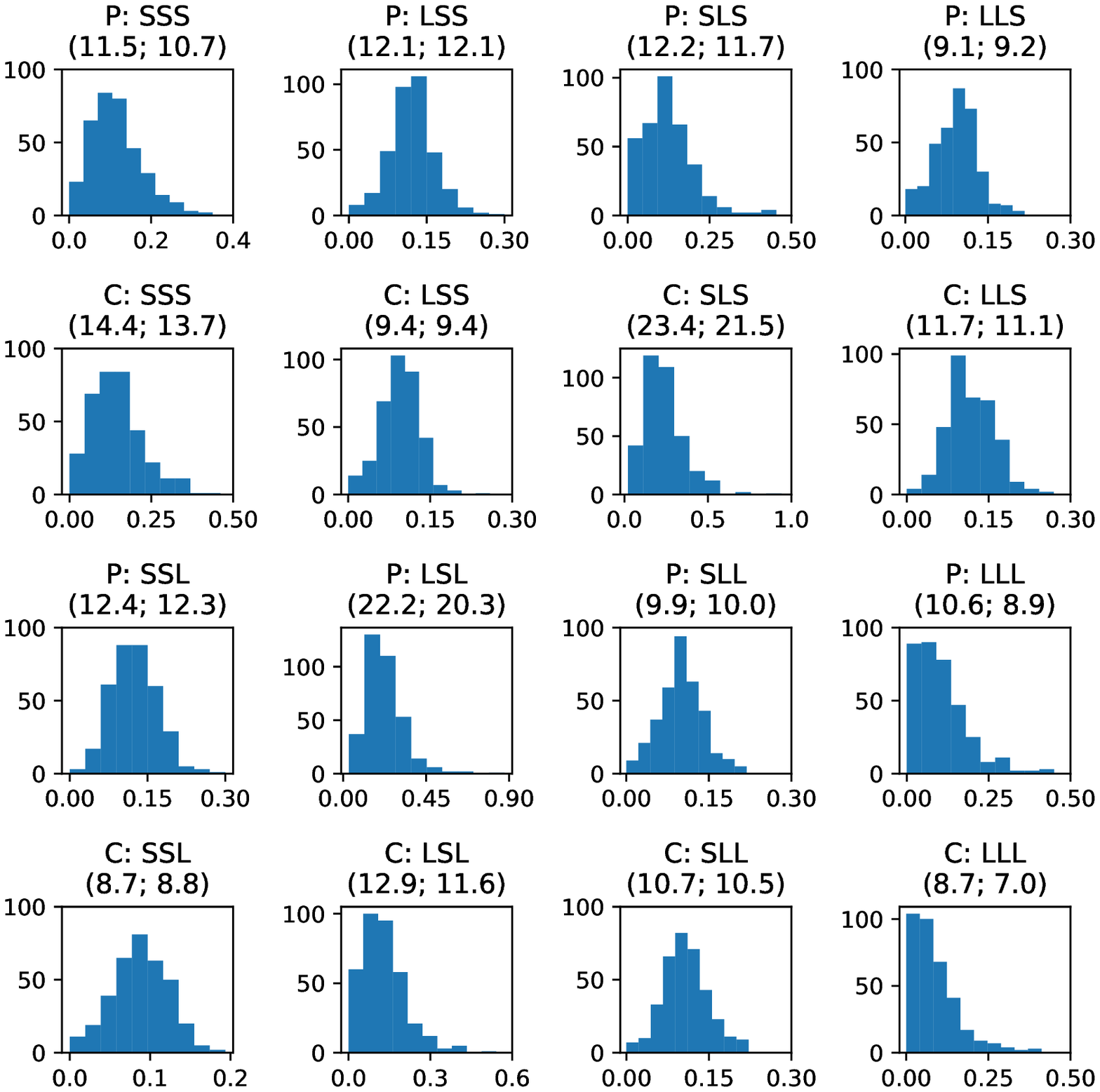}
\caption{\textbf{State distributions in PCCRI corpus.} The distribution of each state in a 3\textsuperscript{rd}-order CODYM, stratified by patient and clinician turns, across all 355 conversations in the PCCRI corpus. Each distribution is labeled by patient (P) or clinician (C) turns, the state, and parenthetically the mean and median values, in that order.  }
\label{fig:normality_check_states}
\end{figure}

\begin{figure}[!h]
\centering
\includegraphics[scale=0.7]{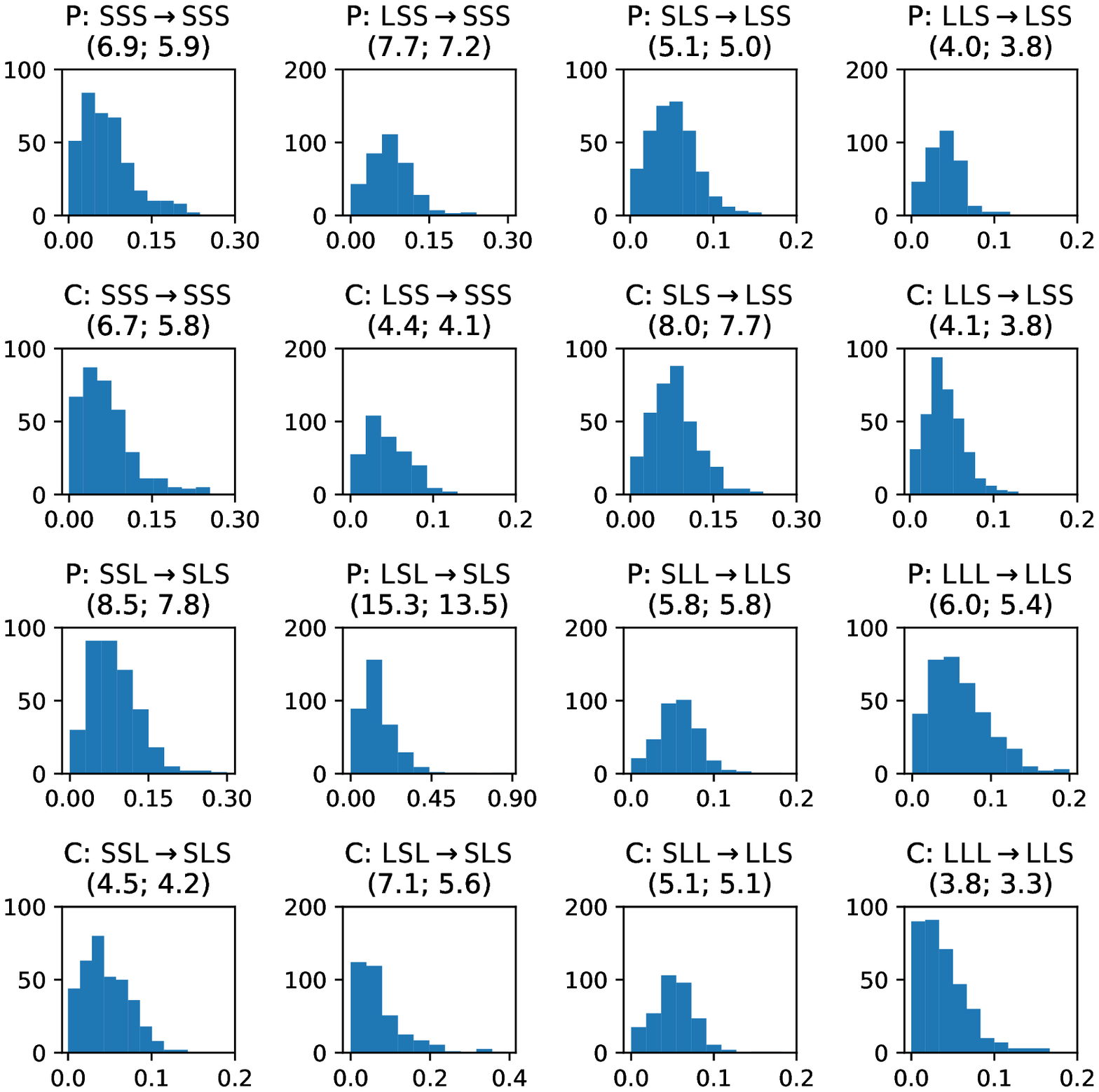}
\caption{\textbf{Transition distributions of short turns in PCCRI corpus.} The distribution of frequencies on each \textbf{short} transition in  3\textsuperscript{rd}-order CODYMs, stratified by patient and clinician turns, across all 355 conversations in the PCCRI corpus. Each distribution is labeled by patient (P) or clinician (C) turns, the transition, and parenthetically the mean and median values, in that order.}
\label{fig:normality_check_trans_S}
\end{figure}

\begin{figure}[!h]
\centering
\includegraphics[scale=0.7]{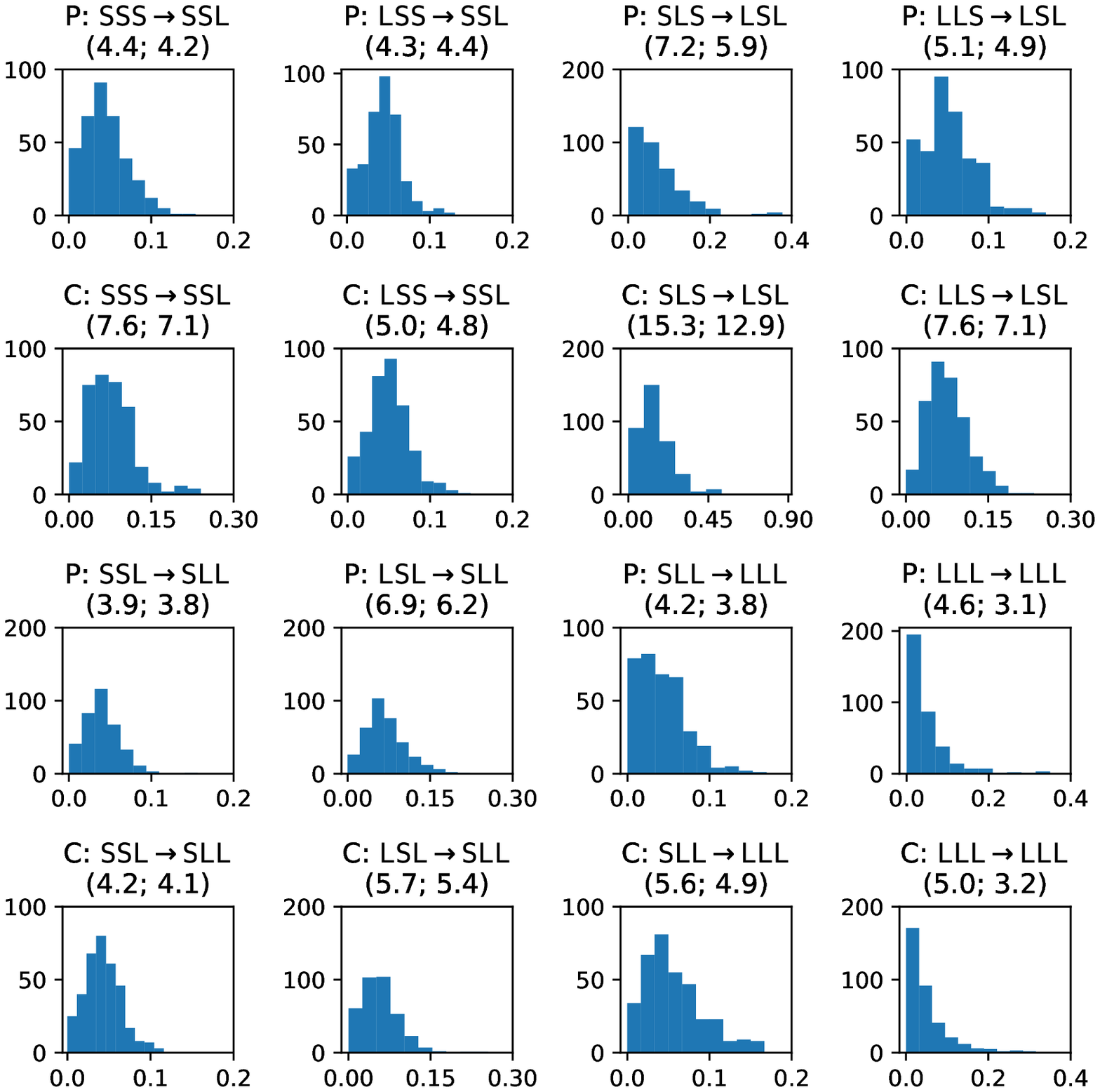}
\caption{\textbf{Transition distributions of long turns in PCCRI corpus.}The distribution of frequencies on each \textbf{long} transition in  3\textsuperscript{rd}-order CODYMs, stratified by patient and clinician turns, across all 355 conversations in the PCCRI corpus. Each distribution is labeled by patient (P) or clinician (C) turns, the transition, and parenthetically the mean/median values.}
\label{fig:normality_check_trans_L}
\end{figure}

CODYMs reveal normative patterns (``fingerprints'') of information flow in serious illness conversations. The existence of such normative patterns is supported by the uni-model distributions observed for all states and transitions in 3\textsuperscript{rd}-order CODYMs, for both patients and clinicians, indicating a prevailing pattern across all conversations in the corpus (Figs. \ref{fig:normality_check_states}, \ref{fig:normality_check_trans_S}, and \ref{fig:normality_check_trans_L}). These patterns are similar, and yet distinct, between patients and clinicians.

When interpreting the populated CODYMs, it is important to consider that only 42\% of patient turns are long whereas 53\% of clinician turns are long. Thus, to create appropriate null models, we randomize the locations of specific turns (thereby preserving the exact distributions of patient and clinician turn lengths), while maintaining the overall sequential order of patient and clinician turn-taking in the actual conversations.  Null 3\textsuperscript{rd}-order CODYMs are generated from 1000 randomized versions of these conversations, stratified by patient and clinician turns. 

Despite the skew apparent in most of these distributions (Figs. \ref{fig:normality_check_states}, \ref{fig:normality_check_trans_S}, and \ref{fig:normality_check_trans_L}), we elect to populate the CODYMs with the means (rather than medians) of transition and state values, since this preserves the more intuitive properties that (a) the sum of all edge weights in a CODYM is 100\%, (b) the sum of all node weights in a CODYM is 100\%, and (c) for each node, the sum of weights of incoming edges equals the sum of the weights of outgoing edges. We verified that using the means, rather than the medians, does not qualitatively change any insights or conclusions drawn from the models (see Figs. \ref{fig:normality_check_states}, \ref{fig:normality_check_trans_S}, and \ref{fig:normality_check_trans_L} for a comparison of mean {\em vs.} median values).

\begin{figure}[!h]
\centering
\includegraphics[scale=0.38]{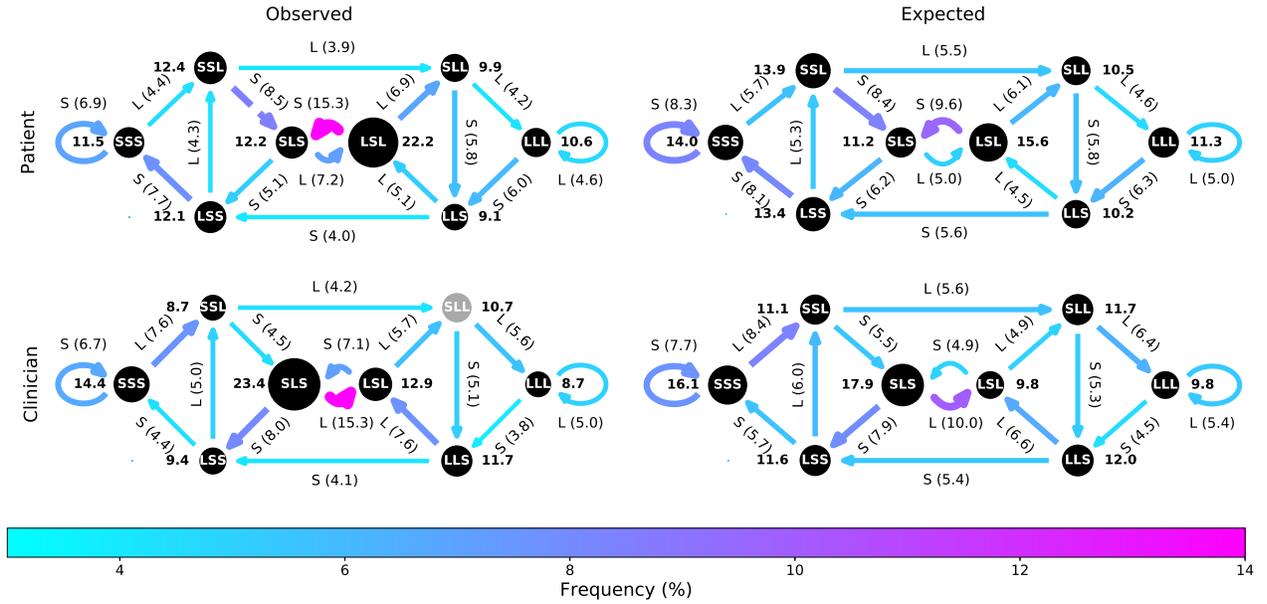}
\caption{\textbf{CODYMs of normative patterns in PCCRI corpus.} CODYMs of normative patterns of information flow for patient turns (top left) and clinician turns (bottom left), averaged over all conversations in the PCCRI corpus. Null models were constructed with the same turn length imbalance for patients and clinicians and the same sequential order of patient and clinician turns in each conversation (right column). Edge thickness and color indicate $\%Observed$ for each transition, as shown parenthetically on edge labels. Node diameter indicates $\%Observed$ for each state, as shown by the node labels in bold. All state and transition values were significantly different from their corresponding null models, according to the empirically derived 95\% confidence intervals, with the exception of state \texttt{SLL} for clinicians (shown in gray).}
\label{fig:normative_pattern2}
\end{figure}

When comparing the observed CODYMs (Fig. \ref{fig:normative_pattern2}, left column) to the expected values from the corresponding null models (Fig. \ref{fig:normative_pattern2}, right column), states \texttt{SLS} and \texttt{LSL}, and the transitions between these two states, occur more frequently in the observed data than expected by chance in both patients and clinicians, whereas all other states occur less frequently than expected by chance (Tables \ref{table:norm_states} and \ref{table:norm_trans}). This indicates that conversations include more one-way information exchanges than would be expected by chance, and that sometimes it is the clinician imparting more information and sometimes it is the patient. 

Despite the qualitative similarities between normative patterns of patients and clinicians described above, all but two of the eight states and five of the sixteen transitions are siginficantly different for the two speaker types (Tables \ref{table:norm_states} and \ref{table:norm_trans}). Most notably, it is evident that the \texttt{LSL} state is the most frequent state prior to a patient turn and is most often followed by a short patient turn (Fig. \ref{fig:normative_pattern2}, upper left) whereas the \texttt{SLS} state is the most frequent state prior to a clinician turn and is most often followed by a long clinician turn (Fig. \ref{fig:normative_pattern2}, lower left). This implies that clinicians, rather than patients, are most often imparting more information in one-way information exchanges.  Other aspects of the normative CODYM patterns are discussed in subsequent sections.
 
\subsection*{Dynamic Changes in Normative Patterns of Information Flow} \label{sec:dynamics}

\begin{figure}[!h]
\centering
\includegraphics[scale=0.36]{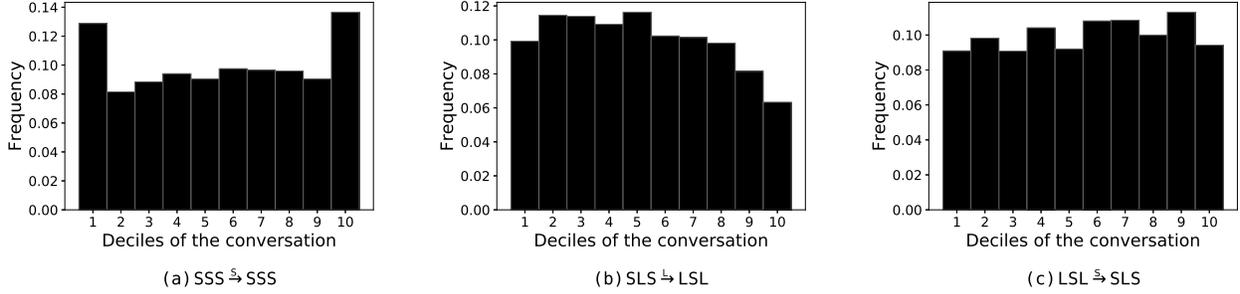}
\caption{\textbf{Selected temporal changes in CODYMs of PCCRI corpus.} Histograms of temporal changes in transition frequencies for patient turns in 3\textsuperscript{rd}-order CODYMs for (a) short two-way information exchanges,  (b) one-way information exchanges from patient-to-clinician, and (c) one-way information exchanges from clinician-to-patient, over 10 conversational deciles that were subsequently averaged over all conversations in the PCCRI corpus and normalized, such that the sum of all bins is 1.0. See Figs. \ref{fig:temporal_plots_S} and \ref{fig:temporal_plots_L} for plots of temporal changes in all transition frequencies, for both patient and clinician turns.}
\label{fig:temporal_plots}
\end{figure}

\begin{figure}[!h]
\centering
\includegraphics[scale=0.7]{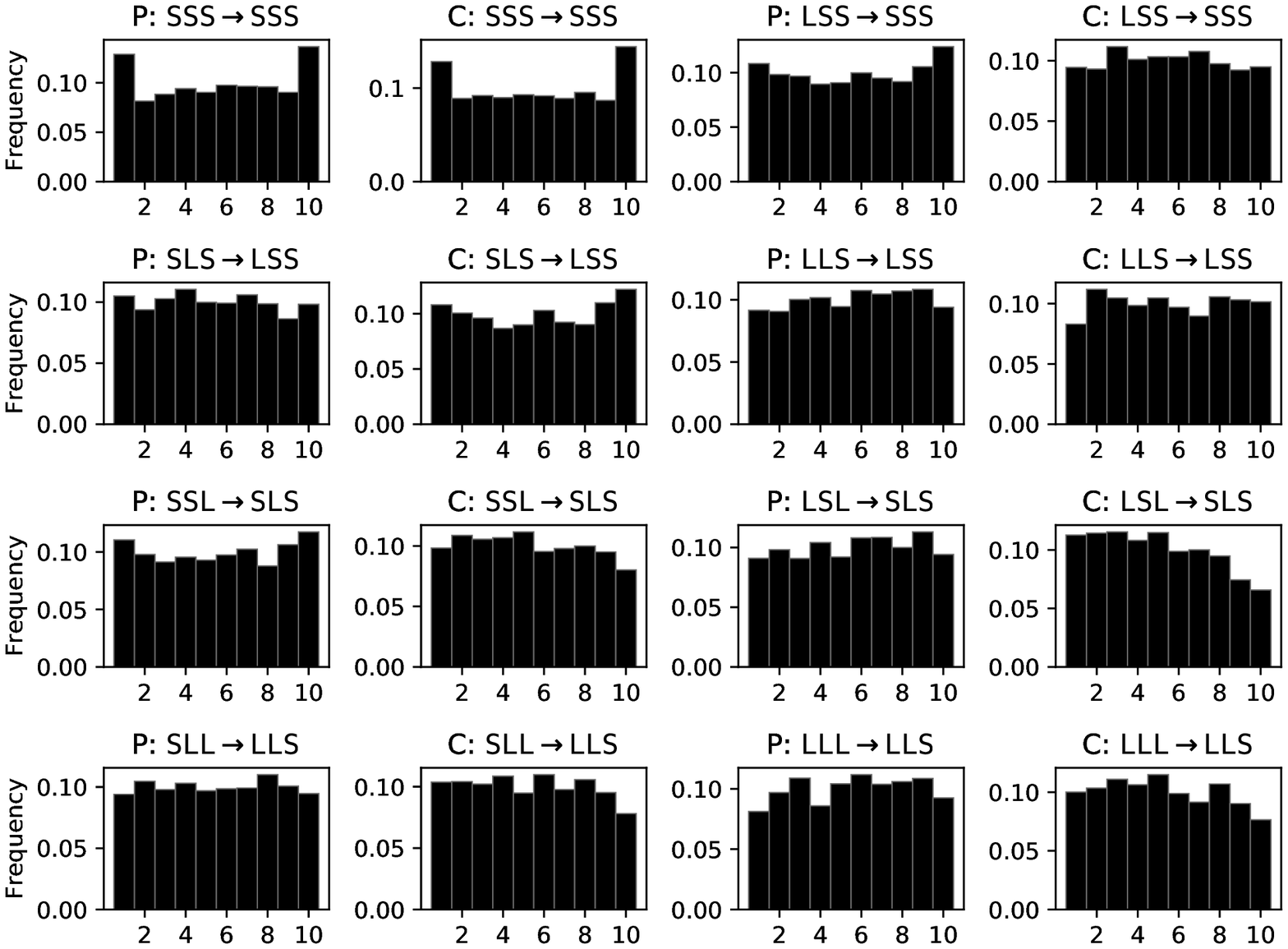}
\caption{\textbf{Temporal changes in transitions of short turns in PCCRI corpus.} Histograms of transition frequencies of all \textbf{short} turns in 3\textsuperscript{rd}-order CODYMs over 10 conversational deciles (normalized, such that the sum of all bins is 1.0), stratified by the patient and clinician turns in the PCCRI corpus.}
\label{fig:temporal_plots_S}
\end{figure}

\begin{figure}[!h]
\centering
\includegraphics[scale=0.7]{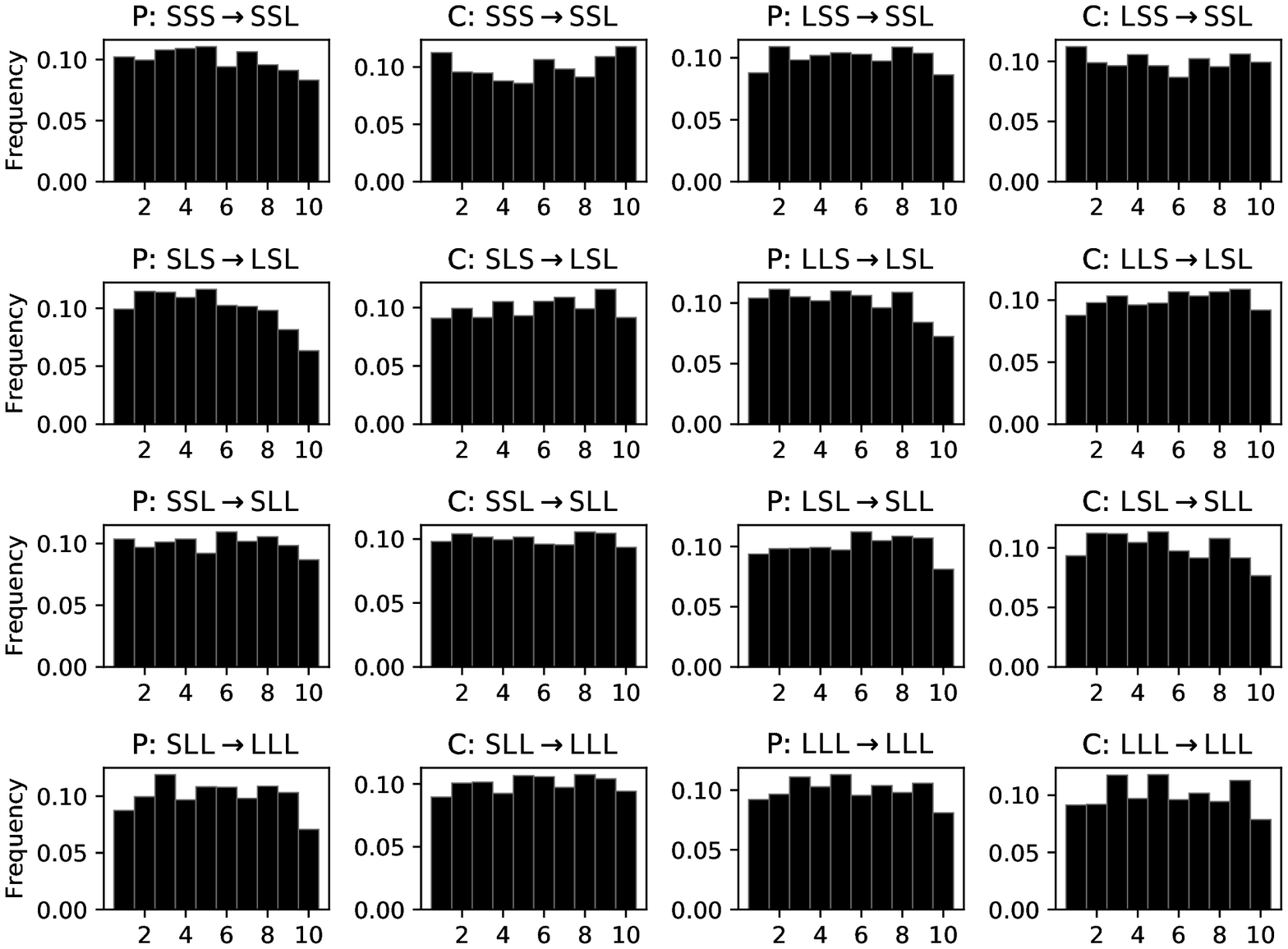}
\caption{\textbf{Temporal changes in transitions of long turns in PCCRI corpus.} Histograms of transition frequencies  of all \textbf{long} turns in 3\textsuperscript{rd}-order CODYMs over 10 conversational deciles (normalized, such that the sum of all bins is 1.0), stratified by the patient and clinician turns.}
\label{fig:temporal_plots_L}
\end{figure}

Some transitions show distinct patterns of change in frequency across temporal deciles of conversations (Figs. \ref{fig:temporal_plots}, \ref{fig:temporal_plots_S}, and \ref{fig:temporal_plots_L}). For example, the short two-way information exchange, while occurring overall less frequently than expected by chance (Fig. \ref{fig:normative_pattern2}), occurs more frequently in the first and last decile of the conversation for both patient turns (Fig. \ref{fig:temporal_plots}a) and clinician turns (Fig. \ref{fig:temporal_plots_S}). This is consistent with our observation that words associated with ritualistic openings and closings of a conversation are over-represented in short two-way information exchanges (see Contextualization by Word Clustering, in the next Section). For patient turns, we also see a decrease in one-way information exchanges from patient-to-clinician from deciles 5 through 10 (Fig. \ref{fig:temporal_plots}b, $\rho=-0.95, p = 0.004$), but no corresponding significant increase in the one-way information flow from clinician-to-patient over deciles 5 through 10 (Fig. \ref{fig:temporal_plots}c, $p = 0.84$). The complementary changes in one-way information flow patterns for clinician turns exhibit nearly identical patterns (Figs. \ref{fig:temporal_plots_S} and \ref{fig:temporal_plots_L}). 

\subsection*{Contextualization} \label{sec:contextualization}

\subsubsection*{Contextualization by word clustering} \label{sec:clusteringcontextualization}

\begin{figure}[h]
\centering
\includegraphics[scale=0.75]{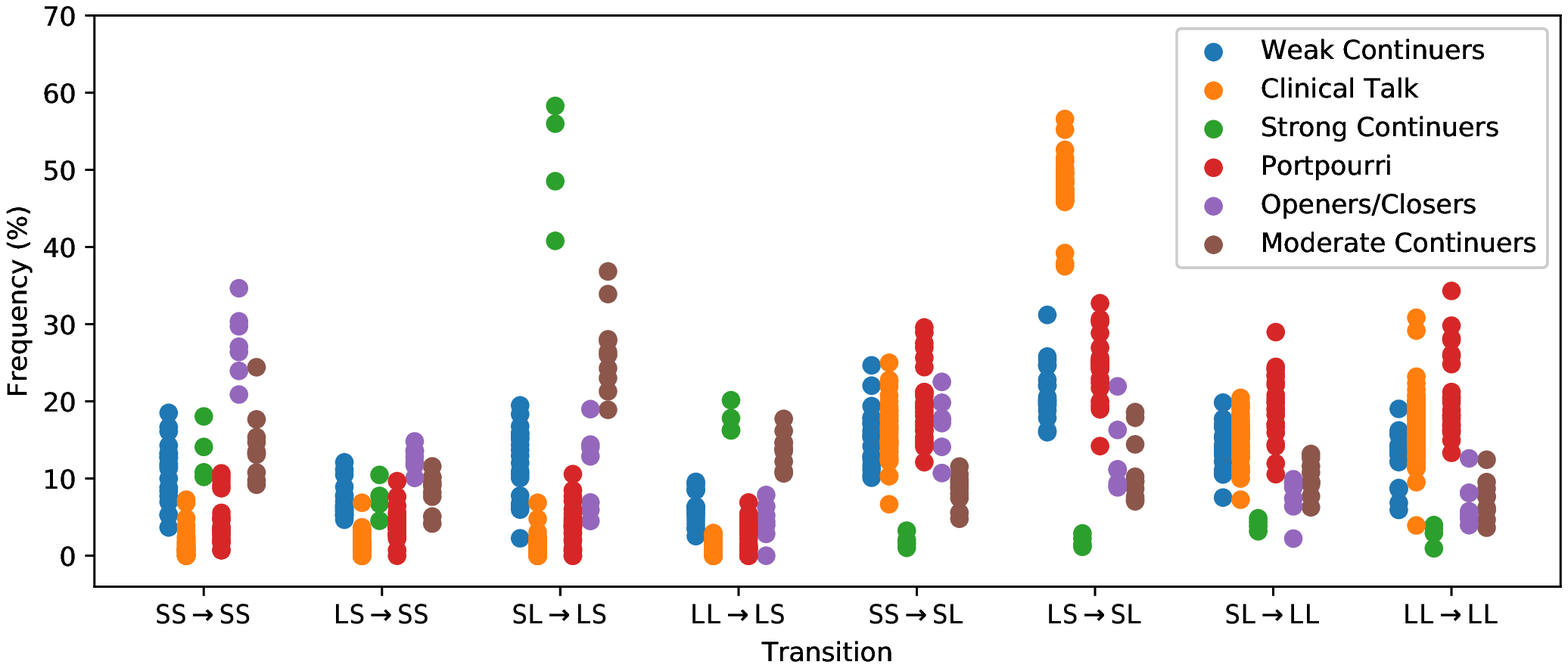}
\caption{\textbf{Distributions of transition frequencies in word clusters of PCCRI corpus.} Transition frequencies, by word cluster, for all words in the PCCRI coprus that are used 100 or more times, and whose transition frequencies differ by at least 10\% from expected. The ordering of clusters (from left to right) in the figure match the ordering of cluster names (from top to bottom) in the legend.}  
\label{fig:cluster_distribution}
\end{figure}

Unsupervised clustering was used to group 114 words (selected as previously described in the Methods) into six clusters, based on the transition frequencies of individual words in 2\textsuperscript{nd}-order CODYMs. For each cluster, we generate a representative CODYM by averaging the transition frequencies for all words in the cluster (for the distribution of transition frequencies in each cluster, see Fig. \ref{fig:cluster_distribution}). We compare these to the CODYM of occurrences of all words (Fig. \ref{fig:clusternull}), to see which transitions are over- or under-represented. Because the CODYM of occurrences of all words is not formulated from a distribution of random trials, it is not possible to empirically derive the confidence interval to determine whether the differences between observed and expected frequencies are significant.  However, the purpose of the clustering is not to determine which state/transition frequencies are significantly different from random, but to observe which words share similar patterns of information sharing.

\begin{figure}[!h]
\centering
\includegraphics[scale=0.4]{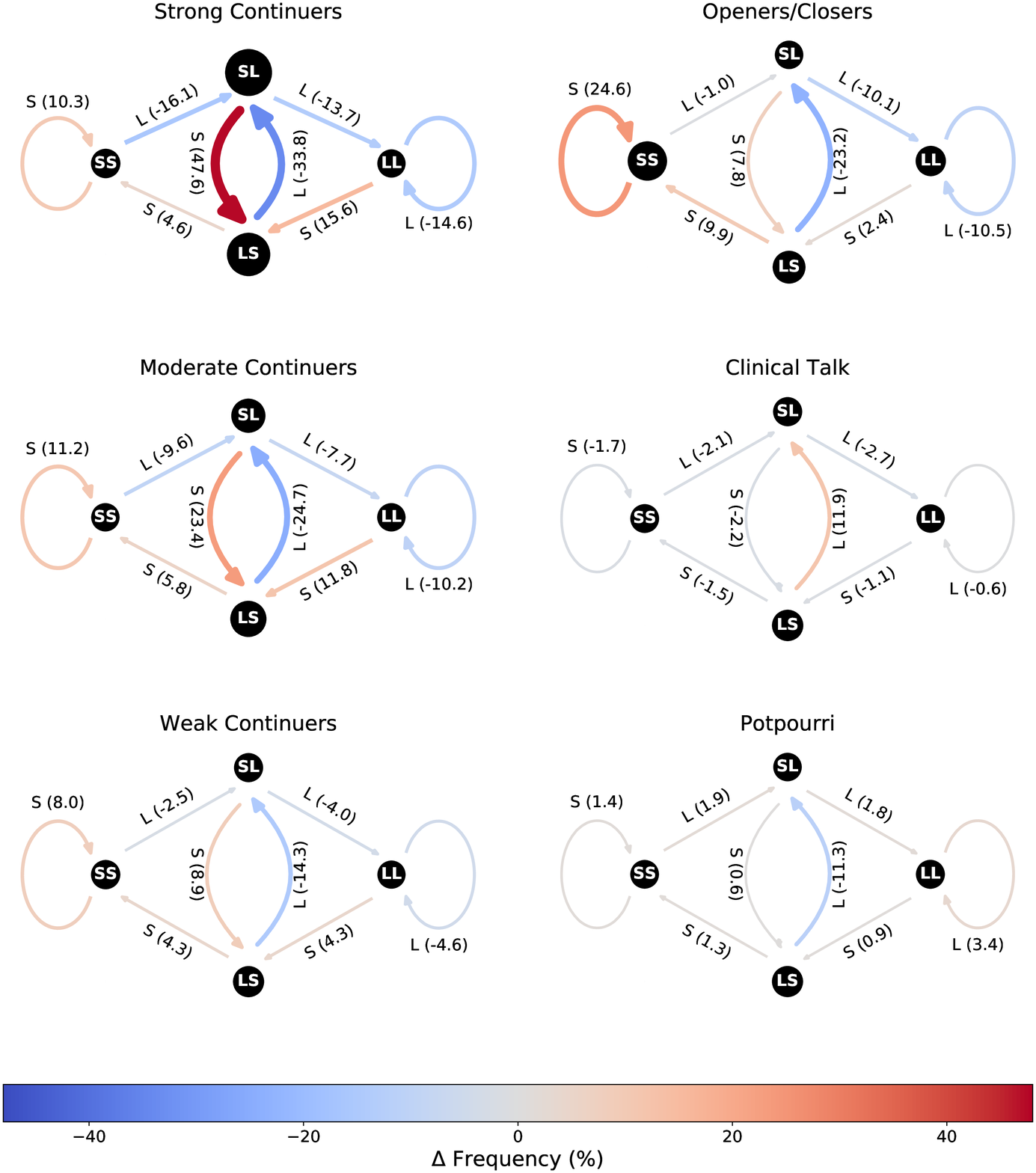}
\caption{\textbf{CODYMs of word clusters in PCCRI corpus.} Differences in  2\textsuperscript{nd}-order CODYM information flow patterns of six clusters of words created by unsupervised clustering (Table \ref{tab:clusters}), relative to expected frequencies based on all words in the PCCRI corpus. Edge thickness, color, and labels indicate the amount by which the frequencies differ from those of all words in the corpus, $\Delta Frequency = \%Observed$ $-$ $\%Expected$, where $\% Expected$ is as shown in Fig. \ref{fig:clusternull}. Node diameter is proportional to $\%Observed$. } 
\label{fig:word_clusters}
\end{figure}

We find that many words with similar information flow patterns fill similar functions in the context of information flow, and we refer to these clusters by monikers according to these functions (Table \ref{tab:clusters}, Fig. \ref{fig:word_clusters}). A complete listing of words in each cluster and their frequencies in the PCCRI corpus are shown in Table \ref{tab:full_word_clusters}. Results are displayed in Fig. \ref{fig:word_clusters} as $\Delta Frequency = \%Observed - \%Expected$ to accentuate how each cluster differs from the norm. (The $\% Observed$ for each transition in each cluster are shown in Table \ref{table:cluster_freqs}.) 

\begin{table}[ht]
\caption{\textbf{Word clusters from CODYMs of PCCRI corpus.}  Six unsupervised clusters of 114 words, based on similarities in transition frequencies in a 2\textsuperscript{nd}-order CODYM. For each cluster, we identify the number of words in the cluster, the percentage of turns that included any words in the cluster that were long (\texttt{L}), and example words in the cluster. See Table \ref{tab:full_word_clusters} for the complete lists of words in each cluster, and how many times each of these words appeared in the PCCRI corpus. }
\begin{tabular}{l|l|l|l} \hline
\textbf{Cluster Moniker} & \textbf{\# Words}  & \textbf{\%\texttt{L}} & \textbf{Example
Words} \\ \hline  % \hhline{|=|=|=|=|}
Strong Continuers &4&10.8& hm, hmmm, mm, aha\\ \hline
Moderate Continuers &10&36.9& yeah, yes, yep, yup, ok, wow, huh\\ \hline
Weak Continuers &19&63.5& fine, right, sorry, no, beautiful, meeting, appreciate\\ \hline
Openers/Closers &7&44.3&thank(s), meet, hi/hello, welcome, bye \\ \hline
Clinical Talk &51&95.5& um, comfort, symptoms, treatments, chemotherapy, continue, disease\\ \hline
Potpourri &23&84.8& hurts, tylenol, concern, risk, confused, scary, funny\\ \hline
\end{tabular}
\label{tab:clusters}
\end{table}

Three similar clusters include words that often occur in \texttt{S} turns during one-way information exchanges, linguistically referred to as ``continuers'' \cite{jefferson1984notes, schegloff1982discourse, yngve1970getting}. We refer to these three clusters as ``strong'', ``moderate'', and ``weak'' continuers, based on the degree to which the one-way information exchange pattern differs from that in the CODYM of all word frequencies. The most specialized grouping, the strong continuer cluster, comprises sounds of acknowledgment that serve to encourage the primary speaker to continue (e.g., hm, aha). These words have the most extreme specialization as continuers and are rarely used in other contexts, nearly always occurring during \texttt{S} turns and, specifically, appearing in the \tran{SL}{S}{LS} transition more frequently than all other transitions combined. The moderate continuer group contains positive affirmations (e.g., yeah, yes, yep, ok) that, while often used as continuers, have a less extreme over-representation in the one-way information exchange, and not infrequently (in over a third of turns) occur in \texttt{L} turns. The weak continuer cluster contains some words that act as continuers in one-way information exchanges (e.g., right, no, nice, great, fine, exactly), but also occur in \texttt{L} turns nearly two thirds of the time and contain some words that aren't generally considered  continuers (e.g., meeting, appreciate, appetite).

The ``openings/closings'' cluster contains words that are almost exclusively associated with starts and ends of conversations, occurring nine times as often in short two-way information exchanges, relative to the CODYM of all word frequencies. This is consistent with our earlier observation that short two-way information exchanges occur most often in the first and last deciles of conversations (Fig. \ref{fig:temporal_plots}). 

The cluster we have dubbed ``clinical talk'' is the largest cluster, containing 51 of the 114 words considered. The words in this cluster most often occur during \texttt{L} turns in one-way information exchanges, with a slight under-representation in all other transitions, relative to the model of all word frequencies.  While many of the words in this cluster specifically relate to clinical talk (e.g., comfort, symptoms, treatments, chemotherapy), others are more general (e.g., sort, may, whether); the most prevalent word in this cluster is ``um'', a word often used for holding the floor during a \texttt{L} turn.

The remaining 23 words were grouped into a cluster we refer to as ``potpourri'', since the relationships between these words are less obvious than in the other clusters. Although several of them described feelings (e.g., hurts, concern, confused, scary, funny), the word risk, a word used predominantly by clinicians (Table \ref{tab:hedgingterms}), also appears in this cluster.  The general pattern of information flow in this cluster is opposite of what was observed for the ``clinical talk'' cluster. Specifically, these words are under-represented on the \texttt{L} turns of one-way information exchanges and slightly over-represented on all the other transitions, more so on other \texttt{L} turns than on \texttt{S} turns, with the greatest over-representation during long two-way exchanges. 

\subsubsection*{Contextualization by hedging and treatment terms} \label{sec:termcontextualization}

\begin{figure}[h]
\centering
\includegraphics[scale=0.42]{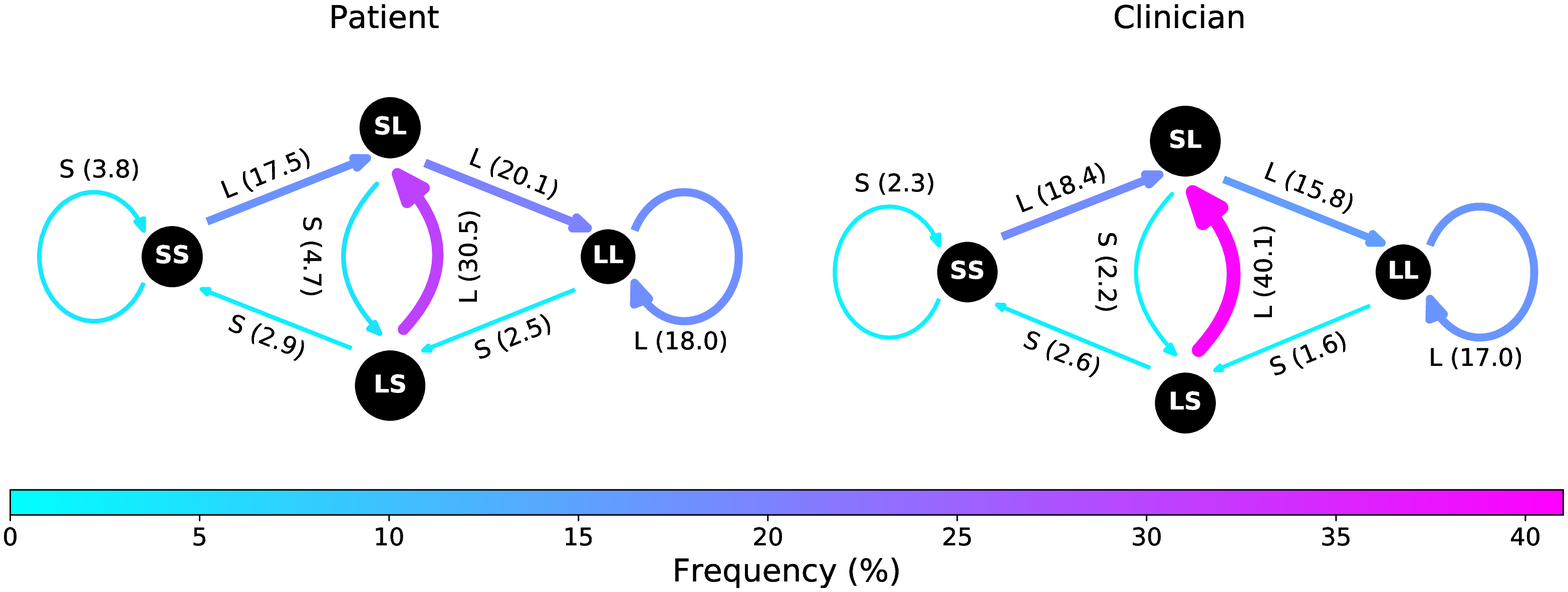}
\caption{\textbf{CODYMs of word usage in stratified PCCRI corpus.} State and transition frequencies of all words in the PCCRI corpus, stratified by patient and clinician turns.  Transition labels indicate the length of the turn on the transition, parenthetically followed by the percentage of word occurrence on that transition. Edge thickness and color indicate $\%Observed$ for each transition, and node diameter indicates $\%Observed$ for each state. 1000 null models were created according to these frequencies to determine expected values for comparison to frequencies of hedging or treatment terms.}  
\label{fig:word_freq_null}
\end{figure}

To assess whether patients and clinicians use hedging and/or treatment terms differently, we stratify the corpus into patient turns and clinician turns. We compare observed CODYM transition frequencies to expected frequencies in null CODYMs, where these null models were created from size-matched samples drawn according to the known frequencies of each state/transition over all patient and clinician turns. In this way, the null models capture the amount of random noise that would be expected due to chance, given the sample size. As shown in Fig. \ref{fig:word_freq_null}, the mean frequencies of these stratified null CODYMs are very similar to the single CODYM of the entire unstratified corpus already shown in Fig. \ref{fig:clusternull}, so are not shown in the main text. 

\begin{figure}[!h]
\centering
\includegraphics[scale=0.4]{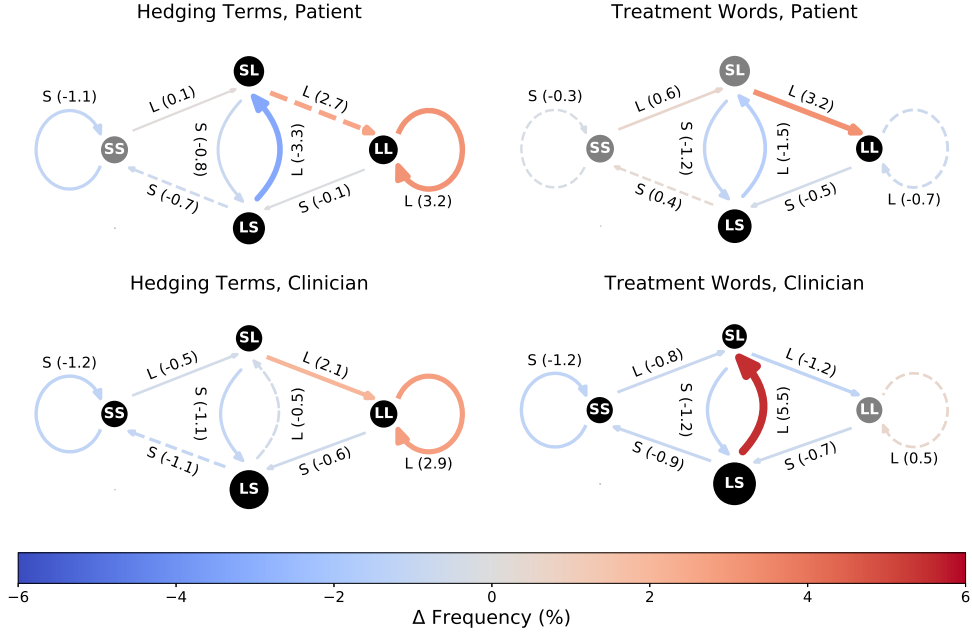}
\caption{\textbf{CODYMs for hedging and treatment terms in PCCRI corpus.} Differences in 2\textsuperscript{nd}-order CODYM patterns for hedging and treatment terms, relative to the means of null models sampled from all words in the corpus, stratified by patient and clinician turns. Edge labels indicate the length of the turn in the transition (\texttt{S} \textit{vs.} \texttt{L}), parenthetically followed by the observed discrepancy in the percentage occurrence. Dashed transition arrows, and nodes colored gray, represent transitions and states where observed frequencies were not significantly different from expected, according to the empirically derived 95\% confidence intervals. Edge thickness, colors, and labels represent $\Delta Frequency = \%Observed - \%Expected$. Node diameter is proportional to $\% Observed.$} 
\label{fig:word_groups}
\end{figure}

\begin{figure}[!h]
\includegraphics[scale=0.37]{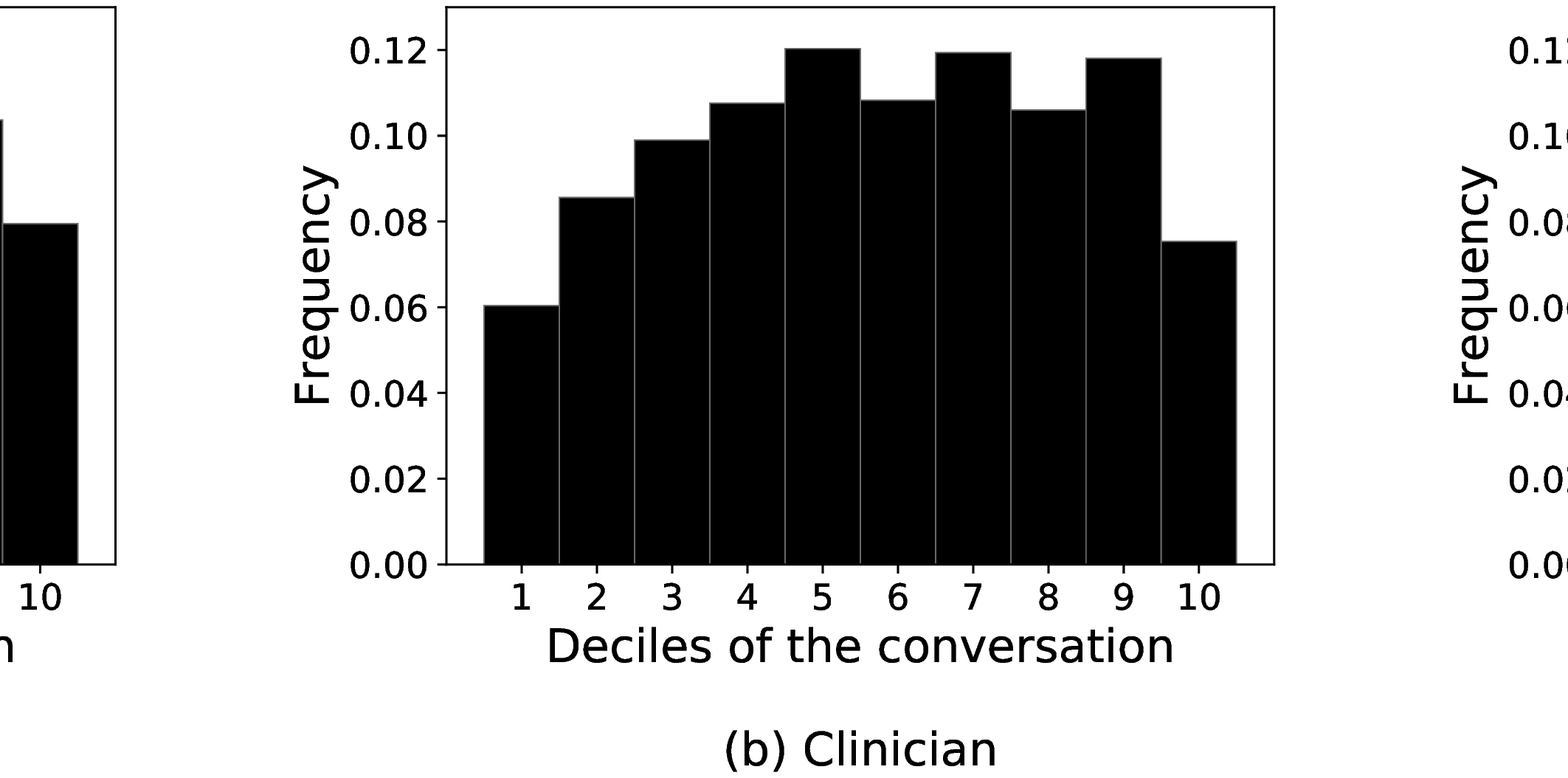}
\caption{\textbf{Temporal use of hedging terms in PCCRI corpus.} Temporal distribution of all turns in the PCCRI corpus that include hedging terms for (a) patients, (b) clinicians, and (c) all speakers across narrative time.}
\label{fig:hedging_temporal}
\end{figure}

Patients and clinicians used hedging terms 3,805 and 5,794  times, respectively. Overall, hedging terms increased in frequency over the first half of the conversation, especially when used by clinicians (for patients, the usage of hedging  terms peaked in decile 3); the use of hedging terms then remained high until decile 9 for both speaker types, dropping in the last decile (Fig. \ref{fig:hedging_temporal}). The differences in CODYM patterns where hedging terms were used, relative to the CODYMs of all words, were similar for patients and clinicians (Figure \ref{fig:word_groups}, left column). Hedging terms are generally used 2-3\% more frequently than expected in \texttt{L} turns following another \texttt{L} turn,  especially during long two-way information exchanges. Patients, but not clinicians, used hedging terms over 3\% less frequently in the  \texttt{L} turn of a one-way information exchange, than in the corresponding null models. All other transitions with significant differences (Fig. \ref{fig:word_groups}, left column, solid transition arrows) had hedging terms slightly under-represented, relative to the corresponding null models. 

Patients and clinicians used treatment terms 4,247 and  10,469 times, respectively. In contrast to the relatively similar CODYM patterns in their use of hedging terms, patients and clinicians exhibited very different CODYM patterns in their use of treatment terms (Figure \ref{fig:word_groups}, right column). Clinicians used treatment terms nearly 6\% more frequently during \texttt{L} turns in one-way information exchanges (from clinician-to-patient), with $\le 1.3\%$ absolute differences from expected for all other transitions, nearly all significantly lower than expected (Figure \ref{fig:word_groups}, lower right, solid transition arrows). Patients, however, used treatment terms over 3\% more frequently than expected on the \tran{SL}{L}{LL} transition, but with $\le 1.3\%$ absolute differences from expected for all other transitions, most of which were not significantly different from the null model (Figure \ref{fig:word_groups}, upper right, dashed transition arrows). 

The overall CODYM patterns shown in Fig. \ref{fig:word_groups} were not qualitatively affected by removal of different random sets of 10\% of hedging terms or treatment terms. This robustness to removal of random terms confirms that the observed patterns within term groupings are not reliant on a particular subset of terms.

\subsubsection*{Contextualization by expression of distressing emotion} \label{sec:emotion}

In total, 208 conversations (58.6\%) contained at least one patient turn where anger, fear or sadness was detected, and in 102 conversations (28.7\%) at least two of these different types of distressing emotion were present. Conversations that had at least one patient turn with a distressing emotion typically (89.4\% of the time) had more than one, with a heavy-tailed distribution indicating that a small number of conversations had very high instances of emotion. It was also common to have multiple consecutive patient turns where emotion was expressed. Consequently, the previous turns, used in defining the state prior to a given transition where emotion was present, also often included expressed emotions. This should be considered when interpreting the results herein.  

\begin{figure}[!h]
\centering
\includegraphics[scale=0.38]{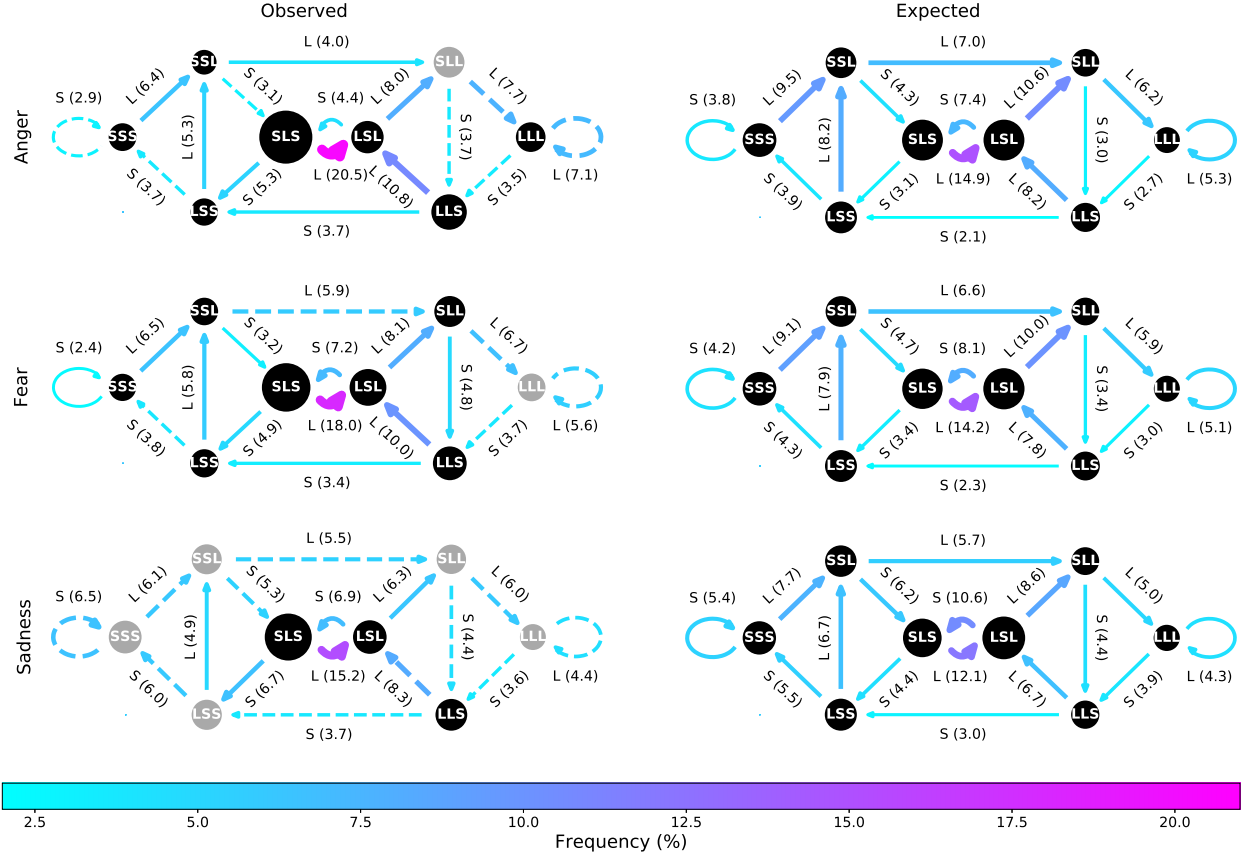}
\caption{\textbf{Turn-level CODYMs by emotional content in PCCRI corpus. } Observed and null CODYMs for patient turns with audibly perceptible expressions of distressing emotion (anger, fear, and sadness) for the PCCRI corpus. Edge thickness and color indicate $\%Observed$ for each transition, as shown parenthetically on edge labels. Node diameter indicates $\%Observed$ for each state. Dashed transition arrows, and nodes colored gray, represent transitions and states where observed frequencies were not significantly from expected, according to the empirically derived 95\% confidence intervals.}
\label{fig:emo_model}
\end{figure}

The expression of anger, fear, and sadness  occurs during \texttt{L} patient turns 69.8\%, 66.6\%, 56.7\% of the time, respectively, whereas only 42\% of all patient turns were \texttt{L} (Fig. \ref{fig:emo_long_rate}). To compare CODYM patterns in patient turns with or without expression of distressing emotions, we analyzed all patient turns with audibly expressed anger, fear or sadness, relative to null CODYMs created by randomly sampling patient turns in which no distressing emotion was present, but with the same number of patient turns and same frequency of \texttt{L} turns as in those patient turns where distressing emotion was detected. As a result of the higher proportion of patient \texttt{L} turns, the resulting null CODYM models for expression of distressing emotion differ from the normative CODYM pattern of discourse for all patient turns (compare the models in Fig. \ref{fig:emo_model}, right column with Fig. \ref{fig:normative_pattern2}, top right).

The observed CODYM patterns during turns where distressing emotion was expressed (Fig. \ref{fig:emo_model}, left column)  differ markedly from what would be expected due to the increased proportion of patient long turns alone (Fig. \ref{fig:emo_model}, right column). Note that, for anger and fear, all but one state and most transitions differ significantly from their respective null model. However, for sadness, only three states and fewer than half of transitions are significantly different from the null model. The most notable differences, present in all three emotions, are (a) an over-representation of expression of distressing emotions during \texttt{L} patient turns in one-way  patient-to-clinician information exchanges, and (b) an under-representation during \texttt{S} patient turns in one-way clinician-to-patient information exchanges. These differences are most extreme for turns in which anger is expressed, and least extreme when sadness is expressed. Additionally, fear is significantly under-represented during short two-way information exchanges, relative to its null model. Expressions of anger, fear, and sadness all peak in deciles 4 and 5 (Fig. \ref{fig:emo_temporal}), and then decrease over the remainder of narrative time in the conversations. The strong similarities in temporal patterns of the three emotions arise, in part, because multiple turns were determined to include more than one of these emotions.

\begin{figure}[!h]
\centering
\includegraphics[scale=0.355]{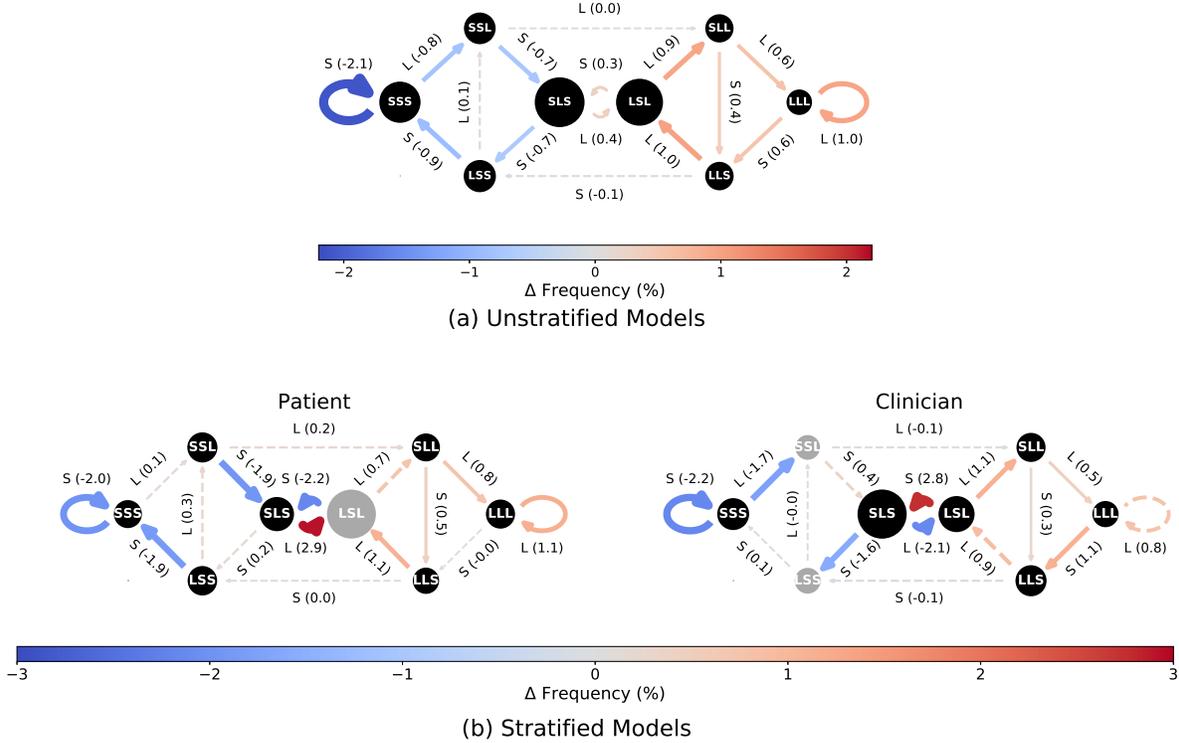}
\caption{\textbf{Conversation-level CODYMs by emotional content in PCCRI corpus. } Differences between  conversations with and without at least one expression of anger or fear, for 3\textsuperscript{rd}-order CODYMs where (a) the data is not stratified by speaker role, and (b) each  conversation is  stratified by patient and clinician turns. Edge labels indicate the length of the turn in the transition (\texttt{S} \textit{vs.} \texttt{L}), parenthetically followed by the observed difference in the percentage occurrence. Dashed transition arrows, and nodes colored gray, represent transitions and states where observed frequencies were not significantly different from expected, according to the empirically derived 95\% confidence intervals. Edge thickness, colors, and labels represent $\Delta Frequency = (\% Observed$ with anger or fear)$-$($\% Observed$ without anger or fear). Node diameter is proportional to $\% Observed$ with anger or fear present.}
\label{fig:emo_conv_combined}
\end{figure}

To examine the impact of distressing emotions on information flow at the conversation level, we focused on  expressions of anger and fear, since expressions of sadness caused relatively few significant differences at the turn level (as seen in Fig. \ref{fig:emo_model}). Out of 355 conversations, 187 (53\%) had at least one instance of anger or fear, creating a relatively balanced data set. The distributions of CODYM transition frequencies for conversations with at least one instance of anger or fear differed significantly from the distributions of CODYM transition frequencies for conversations without any expression of fear or anger, for 13 of 16 transitions (2-sample Kolmogorov-Smirnov test, $p<0.05$, Fig. \ref{fig:emo_conv_combined}a solid edges).  In conversations with at least one expression of  anger or fear, transitions through states with a predominance of \texttt{L} turns, including long two-way information exchanges, and transitions that perpetuate one-way information exchanges, are over-expressed (shown in warm colors in Fig. \ref{fig:emo_conv_combined}a). Conversely, transitions through states dominated by \texttt{S}, especially short two-way information exchanges, are under-expressed (shown in cool colors in Fig. \ref{fig:emo_conv_combined}a). CODYMs from conversations that are stratified by patient and clinician turns tell a more nuanced story.  Not only do these show a larger range in absolute differences of transition frequencies between conversations with and without anger or fear than in the unstratified CODYM model  (note the different ranges on the colorbars in Figs. \ref{fig:emo_conv_combined}a,b), but there are complementary changes in patients and clinicians. Most notably, conversations where anger or fear are expressed have more one-way information flow from patients to clinicians (Fig. \ref{fig:emo_conv_combined}b, left) and less one-way information flow from clinicians to patients (Fig. \ref{fig:emo_conv_combined}b, right), than do conversations with no anger or fear expressed.  In addition, we see that the patient extends long two-way information exchanges significantly more often in conversations where they are expressing anger or fear.

The differences in CODYM transition frequencies between conversations with or without expressions of anger or fear were significant enough to produce a signal with predictive power. Random Forest classifiers, trained on conversation-level transition frequencies of  1\textsuperscript{st}- through 5\textsuperscript{th}-order CODYMs, were able to predict which conversations included any expression of anger and/or fear more often than expected by chance ($p < 0.05$,  Table \ref{table:class_order}). Prediction accuracy (averaged over 1000 trained classifiers)  was (a) slightly higher when using data from stratified {\em vs.} unstratified CODYMs, and (b) generally increased with increasing CODYM order. For example, the 3\textsuperscript{rd}-order stratified transitions frequencies (shown in Fig. \ref{fig:emo_conv_combined}) yielded an average prediction accuracy of 64.5\%, whereas 5\textsuperscript{th}-order stratified transition frequencies raised that to 67.3\%. Even 1\textsuperscript{st}-order stratified transition frequencies exhibited an average prediction accuracy of nearly 60\%.

\section*{Discussion} \label{sec:discussion}

We have shown how CODYM analysis enables one to quantify, visualize, and compare high-level patterns in conversational dynamics from one or more conversations. We have made a deliberate choice to keep CODYMs simple, requiring only sequences of binarized turn lengths and using memory of only 2-3 turns. We have made these choices in order to facilitate interpretability, ensure adequate sample sizes, and to protect privacy. Because of this, there are many important conversational features that CODYMs, as defined here, do not directly incorporate, such as interruptions, overlapping speech, conversational pauses, accent, intonation, gestures, facial expressions and eye contact, all of which play some role in conversational discourse  (e.g., see \cite{durieux2018identifying} for a discussion of the importance of ``connectional silences'' in serious illness conversations). Although any of the aforementioned features {\em could} be directly incorporated into a CODYM, the state space of the resulting model would be larger, less well populated, and more difficult to interpret. As illustrated by our analyses of hedging terms, treatment terms, and expression of distressing emotion, even simple turn-length-based CODYMs provide a means to consider additional verbal and non-verbal features, through contextualization.

CODYM visualizations effectively summarize the overall patterns of sequential dependencies in speaker turn lengths in concise plots, whose size and complexity are independent of the number of turns being analyzed. This facilitates rapid identification and comparison of patterns of information flow in sequences of turns that may include all or parts of one or many conversations.  Since CODYM construction does not require access to detailed conversational content, the approach completely preserves the privacy of conversational participants and can conceptually be incrementally constructed in real-time, precluding the need for storage, or even transcription, of conversations (as discussed further in Conclusions and Future Work). Although CODYMs can be contextualized and can be used to study temporally discretized patterns in information flow, they are not natively designed to capture the turn-level flow of concepts in individual conversations.    CODYMs and Discursis \cite{angus2011conceptual} are thus complementary methods for visualizing information flow in conversations at different scales and levels of detail. 

Through unsupervised word clustering, we discovered that related groups of words often exhibit similar CODYM patterns. This, in itself, helps to validate that CODYMs represent semantically meaningful patterns of information flow.  Three of the emergent clusters contained words that are specialized, to different degrees, for use in the short turns of one-way information exchanges, consisting primarily of strong, moderate, and weak continuers. This is consistent with conversation analytic theory, which suggests continuers are words or phrases used by speakers to convey they are relinquishing an opportunity to command the speaking floor, thus signaling the currently dominating speaker to continue \cite{jefferson1984notes, schegloff1982discourse, yngve1970getting}. Words in the strong and moderate continuer clusters alone appear  44,548 times in the PCCRI corpus (Table \ref{tab:full_word_clusters}), accounting for 41.8\% of all short turns in the corpus. The short two-way information exchange appears less frequently than expected by chance (Fig. \ref{fig:normative_pattern2}), with one notable exception. Word clustering identified a group of openers and closers that occur disproportionately often during short two-way information exchanges (Fig. \ref{fig:word_clusters}).  The importance of openers and closers in conversation was explored by some of the early pioneers of conversation analysis \cite{ schegloff1968sequencing,schegloff1973opening} and has been the subject of much attention since then \cite{miller1975elements, button1990varieties, robinson1998getting, lebaron2002closing, markman2009so}.  These ritualistic transitions into and out of a conversation serve an important role, although little information need actually be exchanged. It is thus to be expected  that openers and closers appear most often in short two-way exchanges (Fig. \ref{fig:word_clusters}), and that this information flow pattern occurs most commonly at the beginnings and ends of conversations (Fig. \ref{fig:temporal_plots}a).

 Conversation has generally been observed to be asymmetric, with a single dominant speaker holding the floor much of the time \cite{itakura2001describing}.  During these one-way information exchanges in didactic conversations, the typical alternating pattern of speakers results in one speaker taking all the long turns (being the ``talker'') while the other takes the short turns (being the ``listener'').  This is particularly true in institutional settings with defined speaker roles \cite{itakura2001describing}, so we expect it will be prevalent in clinical conversations, given the power dynamics between clinicians and patients. Consistent with this expectation, one-way information exchanges from clinician-to-patient were observed to be the most common information sharing pattern in the PCCRI corpus (Fig. \ref{fig:normative_pattern2}), and occurred throughout serious illness conversations at a roughly constant level (Fig. \ref{fig:temporal_plots}c). It is likely that these one-way clinician-to-patient information exchanges  are often used  for delivering information and treatment options related to the patient's illness. Thus, it is not surprising  that treatment terms are most often used when clinicians are in the talker role (Fig. \ref{fig:word_groups}), and words commonly used in clinical talk clustered together as words occurring most often during one-way information exchanges (Fig. \ref{fig:word_clusters}), presumably when the clinician is the talker. 

\begin{figure}[!h]
\centering
\includegraphics[scale=0.5]{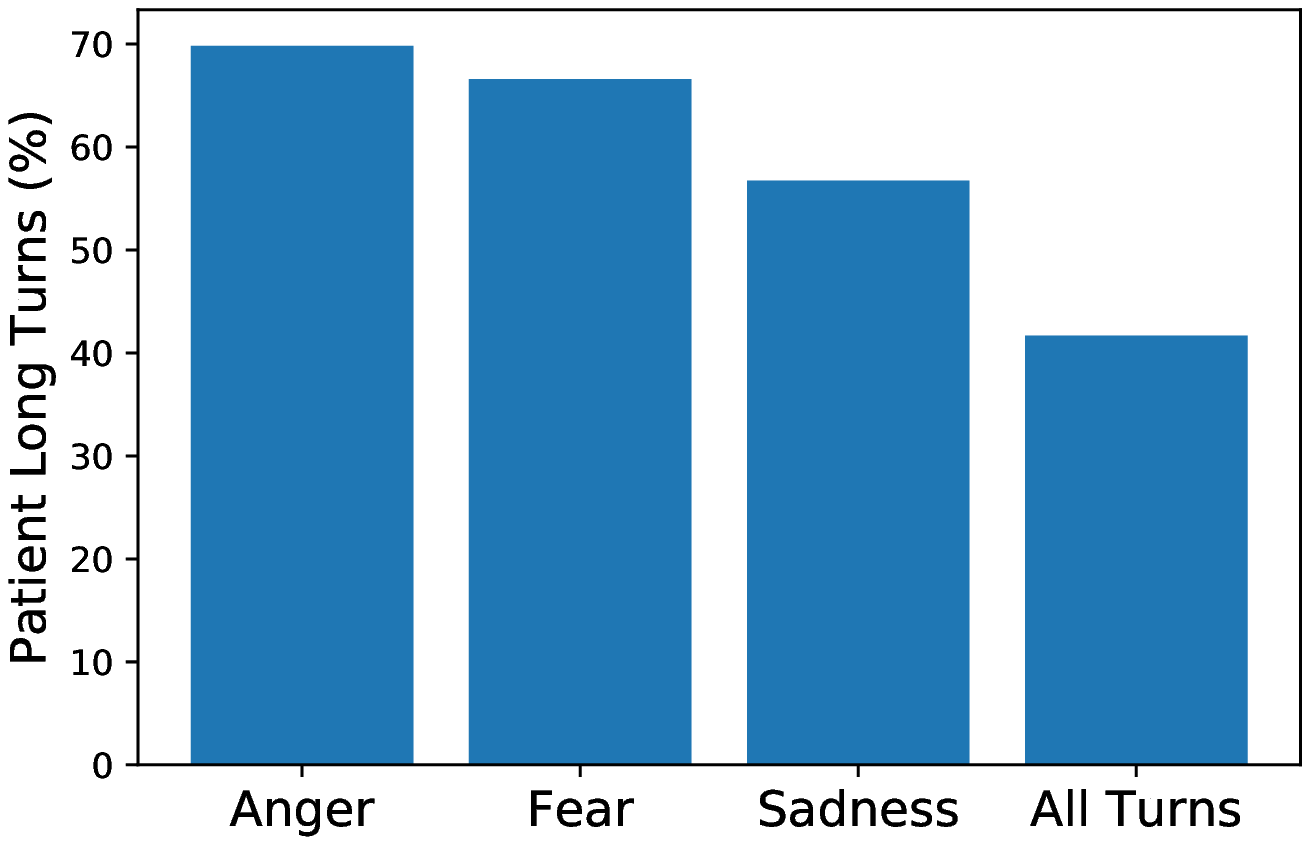}
\caption{\textbf{Turn length by emotion in PCCRI corpus.} Percentage of turns in which distressing emotion (anger, fear, sadness) are expressed that are long, compared to all patient turns.}
\label{fig:emo_long_rate}
\end{figure}

\begin{figure}[!h]
\includegraphics[scale=0.37]{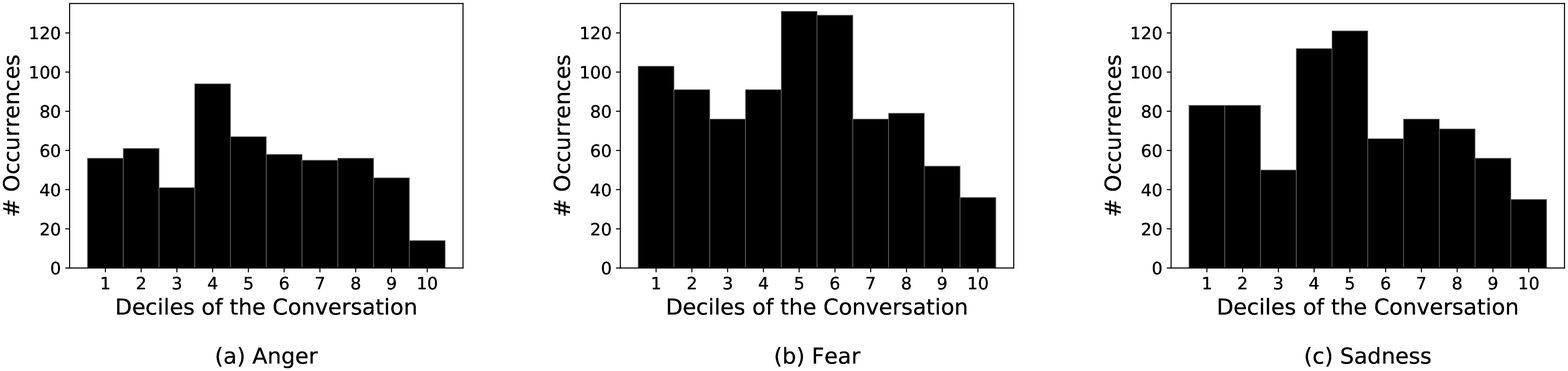}
\caption{\textbf{Temporal expressions of emotion in PCCRI corpus.} Temporal distribution of turns in the PCCRI corpus that include patient expressions of (a) anger, (b) fear, and (c) sadness across narrative time.}
\label{fig:emo_temporal}
\end{figure}

Overall, patients adopt the talker role less often than clinicians do, although still more frequently than expected by chance, in serious illness conversations (Fig. \ref{fig:normative_pattern2}). In pooled patient turns, over all conversations in the corpus, one-way information exchanges from patient-to-clinician are most frequent near the centers of conversations and then steadily decline (Fig. \ref{fig:temporal_plots}b). These temporal changes in patient as talker are similar to the temporal patterns in the usage of  treatment terms reported in \cite{ross2020story}, and also roughly coincide with temporal patterns in the expression of distressing emotions (Fig. \ref{fig:emo_temporal}). We also note that patient expressions of distressing emotion occur disproportionately often in long turns (Fig. \ref{fig:emo_long_rate}), in particular during one-way patient-to-clinician information exchanges (Fig. \ref{fig:emo_model}).  In our previous work, we observed that sentiment scores become increasingly positive over the course of the PCCRI conversations, which was largely attributed to a decrease in the use of disease-related terms that are ascribed negative sentiments \cite{ross2020story}. We hypothesize that the palliative care clinicians in the PCCRI, trained to be highly skilled communicators, may be taking on the role of ``good listener'' \cite{itakura2001describing} during portions of the conversation in which difficult topics are being discussed, as a means of encouraging the patients to express their values and preferences relating to available treatment options. After the patient has finished expressing themselves and one-way patient-to-clinician information exchanges abate, these serious illness conversations may then naturally come to a close.

We also observed an increase in the overall frequency of one-way patient-to-clinician information exchanges, along with a decrease in the  prevalence of short two-way information exchanges and an increase in long two-way information exchanges, in conversations where fear or anger are expressed  (Fig. \ref{fig:emo_conv_combined}).   While these differences at the conversation level are modest, they are strong enough to have predictive power in  distinguishing conversations with anger or fear from those without. This is especially true when the data are stratified by patient and clinician turns; but even when the speaker role is completely anonymized we find that  differences in information flow patterns can be a marker for expression of distressing emotions. Our findings regarding the impact of distressing emotions on patterns of conversational dynamics are important because anger, fear, and sadness manifest frequently in palliative care conversations \cite{alexander2014emotional}, and have been found to have therapeutic effects.  Being able to process and express distressing emotions is linked to improved health outcomes for patients with cancer \cite{stanton2000emotionally}, and it has recently been shown in the PCCRI corpus that expression of anger, in particular, is associated with improvement in how much patients feel heard and understood by the clinical team following a palliative care consult \cite{gramling2020epidemiology}.\\

Uncertainty is prevalent in the discussion of prognosis \cite{sullivan2007diagnosing} and treatment options \cite{mccormack2011measuring}, where outcomes are often unknown. Effectively communicating this uncertainty is essential to having patients fully informed before they make decisions regarding their care \cite{epstein2004communicating, chow2018s}. Helping patients to manage uncertainty is fundamental to patient-centered communication and may be a key ingredient in the beneficial effects of palliative care consultations on patient quality of life \cite{mccormack2011measuring, kimbell2015managing,gramling2018distress}. Hedging terms are often used to soften claims and insert uncertainty into a statement, and are the most prevalent subtype of uncertainty found in the PCCRI corpus \cite{durieux_2019}. As we found previously, the general story arc of conversations in the PCCRI corpus moves from discussion of symptoms, to treatments, to prognosis, to use of modal verbs, peaking in deciles 2,4,6,9, respectively \cite{ross2020story}. Here, we observed that the use of hedging terms by clinicians is highest from deciles 5-9, coinciding with the parts of the conversations where prognosis terms and modal verbs (indicators of possibility) are most frequent. This suggests that clinicians may be using hedging terms in conveying prognostic uncertainty. 

Long two-way information exchanges represent conversational regimes where alternating speakers are trading information.  Although this is one of the least frequent information sharing patterns in the corpus (Fig. \ref{fig:normative_pattern2}), it is more frequent in conversations where audible patient expressions of anger and fear occur (Fig. \ref{fig:emo_conv_combined}) and in turns when commonly used words associated with feelings are expressed (Fig. \ref{fig:word_clusters}). It is thus interesting that long two-way exchanges are also over-represented when hedging terms are used, for both patients and clinicians (Fig. \ref{fig:word_groups}). This may indicate that there is bilateral sharing regarding uncertainty about difficult topics, as would be expected to occur when clinicians are actively helping patients to manage their uncertainty.

\section*{Conclusions and future work} \label{sec:conclusion}

We have presented and validated a novel approach to  quantify and visualize the dynamics of information flow in conversations with CODYMs (COnversational DYnamics Models). CODYMs are the first Markov Model to use speaker turn length as the fundamental unit of information and to provide concise, high-level, quantitative summaries of overall dependencies in sequences of speaker turn lengths. This new approach facilitates identification and comparison of normative patterns of information flow across sequences of turns from one or more conversations, in context-independent or context-dependent ways. CODYMs complement existing qualitative and quantitative approaches for studying conversational dynamics, and comprise a new tool for conversational analysis. We provide open source code for populating, visualizing, and contextualizing CODYMs \cite{CODYMcode}.

We applied the method to a unique and important corpus of palliative care consultations with seriously ill patients. We discovered normative patterns of information flow in these conversations that differ between patients and clinicians, and between conversations with and without expressions of distressing emotions. While these normative patterns are interesting in their own right, they may also have practical applications. For example, it would be interesting to compare normative CODYMs from in-person palliative care consultations to those conducted remotely, to see how telehealth platforms impact the patterns of information flow between patients and clinicians. Similarly, CODYM analysis of mock consultations between clinicians-in-training and actors portraying seriously ill patients could be used to assess how closely the information flow patterns of these training scenarios reflect those observed in palliative care consultations with real patients. If associations can be found between CODYM patterns and quality indicators of healthcare conversations (e.g., the degree to which patients feel heard and understood \cite{gramling2016feeling}, a measure that is currently being considered for widespread use \cite{MACRA2019}), these could provide valuable insights for institutions seeking to improve the quality of conversations with seriously ill patients. 

Achieving patient-centered care, in which clinical decisions are guided by patient preferences, needs, and perspectives \cite{catalyst2017patient}, has long been recognized as a means for improving healthcare delivery \cite{baker2001crossing}.  Patient-centered communication, which includes recognizing and responding to emotions, helping patients to manage uncertainty, promoting reciprocal exchanges of information to create a shared understanding, and helping patients make informed decisions regarding their care, is central to achieving this aim \cite{epstein2007patient}. However, it is not clear how to effectively measure or assess patient centered communication \cite{mccormack2011measuring}. In this work, we have shown how CODYM analysis offers a new approach to quantifying and visualizing patterns of information flow related to markers of patient centered communication, such as the rates of one-way patient-to-clinician information exchanges and long two-way information exchanges, and how these rates change when distressing emotions or uncertainty are expressed. CODYMs may thus be helpful in assessing the degree to which clinical conversations are patient centered.  For example, such an analysis could potentially be useful in understanding observed racial and ethnic disparities in prognosis communication  \cite{ingersoll2019racial}. 

It is not clear whether the conversational ``fingerprints'' uncovered in the PCCRI corpus are unique to serious illness conversations, or represent more general conversational paradigms in healthcare or other contexts. We suspect that the frequency of one-way patient-to-clinician information exchanges, and long two-way information exchanges, may be higher in serious illness communication relative to conversations in other clinical contexts  (e.g., \cite{chhabra2013physician}). It will be fascinating to compare CODYMs across a wide variety of corpora from different languages, cultures, and contexts (including online conversations), to  reveal which  patterns of information flow in conversations are universal, and which are unique to certain settings. 

Conversation analysis has traditionally been a discipline reliant on manual transcription of conversations with highly detailed annotations \cite{hutchby2008conversation}. This is a resource-intensive process that requires full access to the often very private content of conversations. Indeed, the CODYM analyses presented here used transcriptions of audio-recordings of sensitive serious illness conversations. However, we envision an alternate formulation of CODYMs that uses turn duration (in seconds), in lieu of the number of words, for defining turn lengths.  A time-based definition of turn length would facilitate real-time automation and analysis of  conversational dynamics, precluding the need for transcription or even storage of conversational audio, thus completely protecting privacy.  Large numbers of conversations are already taking place in a medium that is natively capable of capturing conversational data appropriate for automated CODYM  analysis. For example, many popular video conferencing services already incorporate tools that automate the detection of speaker turns, and such services have exploded in popularity in the wake of the Covid-19 global pandemic. Ongoing advances in the automated detection of conversational features including speaker recognition \cite{gomez2019designI,gomez2019designII}, emotion \cite{schuller2018speech, zhang2019modeling}, conversational pauses \cite{durieux2018identifying}, empathy \cite{alam2018annotating,chen2020automated}, gaze patterns \cite{sogo2013gazeparser}, and word recognition \cite{tabibian2020survey}, will facilitate real-time analysis and contextualization of CODYMs. Ultimately, we foresee a fully-automated pipeline for CODYM analyses, with no compromise to the privacy of conversational content. 

As more conversational data become available, whether as transcriptions or through real-time processing, CODYMs will be a valuable tool for studying information flow in a wide variety of contexts and contributing to our understanding of how to have more effective conversations. Such a tool could be of  practical utility in training and assessment of high quality communication in healthcare and other application domains, while also yielding new theoretical insights into conversational dynamics across languages, cultures, and contexts.

\section*{Acknowledgments}
This work was funded in part by a Research Scholar Grant from the American Cancer Society (RSG PCSM124655; PI: Robert Gramling), by the Holly \& Bob Miller Endowed Chair in Palliative Medicine at the University of Vermont Larner College of Medicine, and by the National Science Foundation (NSF OIA 1556770; PI: Arne Bomblies, and NSF CRISP 1735513; PI: Paul Hines). We thank the American Cancer Society and the palliative care clinicians, patients, and families who participated in this work for their dedication to enhancing care for people with serious illness. We also thank S.B. Heinrich for constructive comments that helped us to improve the manuscript.

\bibliographystyle{unsrt}  

%\begin{thebibliography}{1}
%\bibliography{references}
%\end{thebibliography}

\newpage
\beginsupplement
\section*{Supplemental Figures and Tables}

\begin{table}[!h]
\caption{\textbf{Classification using CODYMs from emotional content in PCCRI corpus. } Classification accuracy of trained Random Forests for predicting which conversations contained at least one instance of anger or fear. Classifiers were trained using transition frequencies of each conversation as input features, for CODYMs of orders one through five. For each model order, we show the number of input features, and the mean  ($\mu$) and the standard deviation ($\sigma$) of the \% classification accuracy, averaged over 1000 trained classifiers, both for data unstratified by speaker type, and for stratified data (where CODYMs were populated by patient and clinician turns, separately, for each conversation).  $P$-values are calculated from Z-scores based on $\mu$ and $\sigma$, relative to the null hypothesis that the prediction accuracy is not better than random (i.e., $\le 50\%$).}
%https://journals.plos.org/plosone/s/tables
 \begin{tabular}{c|c|c|c|c|c|c|c|c} 
 \hline
 {} & \multicolumn{4}{c}{Unstratified Data} & \multicolumn{4}{|c}{Stratified Data}\\ \hline
 \textbf{Order} & \textbf{\# of Features} & \textbf{$\mu$ (\%)} & \textbf{$\sigma$ (\%)} & $p$ & \textbf{\# of Features} & \textbf{$\mu$ (\%)} & \textbf{$\sigma$ (\%)} & $p$ \\ [0.5ex]
 
 \hline
 1 & 4 & 58.8 & 5.3 & 0.048 & 8 & 59.5 & 5.5 & 0.042\\ 
 \hline
 2 & 8 & 60.3 & 5.2 & 0.024 & 16 & 62.4 & 5.5 & 0.012\\
 \hline
 3 & 16 & 64.4 & 5.5  & 0.004 &32 & 64.5 & 5.3 & 0.003\\
 \hline
 4 & 32 & 64.2 & 5.2 & 0.003 & 64 & 64.8 & 5.6 & 0.004\\
 \hline
  5 & 64 & 65.4 & 5.3 & 0.002 &128&67.3& 5.2 & < 0.001\\
 \hline
\end{tabular}

\label{table:class_order}
\end{table}

\begin{table}[!h]
\caption{\textbf{Hedging terms.} All ``hedging terms'' (from \cite{durieux_2019}) along with the number of times each term was used by patients and clinicians in the PCCRI corpus. Note that terms ending in a ``*'' represent roots and match any word that begins with this term.  In addition, ``allude to'' is a 2-word term.}
 \begin{tabular}[t]{l|r|r} \hline
\textbf{Term}     & \textbf{Patient} & \textbf{Clinician} \\ \hline % \hhline{|=|=|=|}
think & 2652 & 4204 \\ \hline
guess & 453 & 199 \\ \hline
worr* & 235 & 316 \\ \hline
hope & 162 & 293 \\ \hline
expect & 79 & 163 \\ \hline
seem & 78 & 159 \\ \hline
imagine & 41 & 108 \\ \hline
consider & 30 & 90 \\ \hline
risk & 21 & 115 \\ \hline
doubt & 18 & 8 \\ \hline
suppose & 16 & 7 \\ \hline
predict & 9 & 24 \\ \hline
suggest & 7 & 59 \\ \hline
theor* & 7 & 9 \\ \hline
estimate & 5 & 14 \\ \hline
foresee & 3 & 4 \\ \hline
anticipate & 2 & 42 \\ \hline
assess & 2 & 17 \\ \hline
perceive & 2 & 4 \\ \hline
contemplate & 1 & 2 \\ \hline
presume & 1 & 1 \\ \hline
ponder & 1 & 0 \\ \hline
suspect & 0 & 18 \\ \hline
hint & 0 & 2 \\ \hline
speculate & 0 & 1 \\ \hline
imply & 0 & 1 \\ \hline
hypothesi* & 0 & 1 \\ \hline
prognosticate & 0 & 0 \\ \hline
presuppose & 0 & 0 \\ \hline
postulate & 0 & 0 \\ \hline
misjudge & 0 & 0 \\ \hline
misinterpret & 0 & 0 \\ \hline
infer & 0 & 0 \\ \hline
deem & 0 & 0 \\ \hline
allude to & 0 & 0 \\ \hline
\end{tabular} 
\label{tab:hedgingterms}
\end{table}

\newpage
\begin{table}[!h]
\caption{\textbf{Treatment terms.} All ``treatment terms'' considered (from \cite{ross2020story}), with the number of times each term was used by patients and clinicians in the PCCRI corpus.}
\begin{tabular}[t]{l|r|r} \hline
\textbf{Term}     & \textbf{Patient} & \textbf{Clinician} \\ \hline 
hospice & 413 & 1435 \\ \hline
chemo & 369 & 157 \\ \hline
radiation & 303 & 467 \\ \hline
medicine & 289 & 816 \\ \hline
surgery & 236 & 199 \\ \hline
medication & 195 & 459 \\ \hline
dilaudid & 186 & 363 \\ \hline
treatment & 181 & 586 \\ \hline
pills & 161 & 177 \\ \hline
tube & 121 & 380 \\ \hline
pill & 120 & 114 \\ \hline
morphine & 111 & 212 \\ \hline
medical & 107 & 455 \\ \hline
oxycodone & 98 & 192 \\ \hline
therapy & 96 & 166 \\ \hline
iv & 93 & 304 \\ \hline
tylenol & 85 & 83 \\ \hline
drug & 83 & 45 \\ \hline
oxygen & 80 & 81 \\ \hline
drugs & 80 & 38 \\ \hline
medications & 79 & 425 \\ \hline
dialysis & 78 & 154 \\ \hline
patch & 73 & 214 \\ \hline
procedure & 72 & 75 \\ \hline
dose & 66 & 430 \\ \hline
meds & 65 & 46 \\ \hline
fentanyl & 64 & 138 \\ \hline
methadone & 63 & 200 \\ \hline
line & 63 & 66 \\ \hline
chemotherapy & 58 & 350 \\ \hline
machine & 56 & 126 \\ \hline
treatments & 54 & 469 \\ \hline
ativan & 48 & 67 \\ \hline
milligrams & 47 & 90 \\ \hline
liquid & 47 & 76 \\ \hline
button & 45 & 116 \\ \hline
medicines & 43 & 402 \\ \hline
ventilator & 42 & 159 \\ \hline
trial & 40 & 74 \\ \hline
feeding & 40 & 67 \\ \hline
management & 39 & 217 \\ \hline
treat & 38 & 191 \\ \hline
icu & 37 & 70 \\ \hline
fluids & 34 & 83 \\ \hline
antibiotics & 33 & 132 \\ \hline
oral & 21 & 82 \\ \hline
nutrition & 20 & 107 \\ \hline
doses & 17 & 135 \\ \hline
resuscitation & 17 & 101 \\ \hline
cpr & 17 & 78 \\ \hline
milligram & 9 & 33 \\ \hline

\end{tabular}
\label{tab:treatmentterms}
\end{table}

\clearpage

\clearpage

\clearpage

\begin{table}[h]
\caption{\textbf{Significance tests of state distributions in PCCRI corpus.} P-values comparing the state distributions of 3\textsuperscript{rd}-order CODYMs in the PCCRI corpus, stratified by patient and clinician (shown in Fig. \ref{fig:normality_check_states}), of observed patient {\em vs.} observed clinician, observed patient {\em vs.} null patient, and observed clinician {\em vs.} null clinician models, using a 2-sample Kolmogorov–Smirnov Test. }
\begin{tabular}{c|c|c|c} \hline
\textbf{State} & \textbf{Patient {\em vs.} Clinician} & \textbf{Patient  {\em vs.} Null} & \textbf{Clinician  {\em vs.} Null} \\ \hline %\hhline{|=|=|=|=|}
\texttt{SSS} & <0.001 & <0.001 & <0.001 \\ \hline
\texttt{SSL} & <0.001 & <0.001 & <0.001 \\ \hline
\texttt{SLS} & <0.001 & <0.001 & <0.001 \\ \hline
\texttt{SLL} & <0.001 & <0.001 & <0.001 \\ \hline
\texttt{LSS} & <0.001 & <0.001 & <0.001 \\ \hline
\texttt{LSL} & <0.001 & <0.001 & <0.001 \\ \hline
\texttt{LLS} & 0.063 & <0.001 & <0.001 \\ \hline
\texttt{LLL} & 0.002 & <0.001 & <0.001 \\ \hline
\end{tabular}

\label{table:norm_states}
\end{table}

\begin{table}[h]
\caption{\textbf{Significance tests of transition distributions in PCCRI corpus.} P-values comparing the transition distributions of 3\textsuperscript{rd}-order CODYMs, stratified by patient and clinician (shown in Figs. \ref{fig:normality_check_trans_S} and \ref{fig:normality_check_trans_L}), of observed patient {\em vs.} observed clinician, observed patient {\em vs.} null patient, and observed clinician {\em vs.} null clinician models using a 2-sample Kolmogorov–Smirnov Test.}
\begin{tabular}{c|c|c|c} \hline
\textbf{Transition} & \textbf{Patient {\em vs.} Clinician} & \textbf{Patient  {\em vs.} Null} & \textbf{Clinician  {\em vs.} Null} \\
\hline 
\tran{SSS}{S}{SSS} & 0.811 & <0.001 & <0.001 \\ \hline
\tran{SSS}{L}{SSL} & <0.001 & <0.001 & <0.001 \\ \hline
\tran{LSS}{S}{SSS} & <0.001 & <0.001 & <0.001 \\ \hline
\tran{LSS}{L}{SSL} & <0.001 & <0.001 & <0.001 \\ \hline
\tran{SLS}{S}{LSS} & <0.001 & <0.001 & <0.001 \\ \hline
\tran{SLS}{L}{LSL} & <0.001 & <0.001 & <0.001 \\ \hline
\tran{LLS}{S}{LSS} & 0.504 & <0.001 & <0.001 \\ \hline
\tran{LLS}{L}{LSL} & <0.001 & <0.001 & <0.001 \\ \hline
\tran{SSL}{S}{SLS} & <0.001 & <0.001 & <0.001 \\ \hline
\tran{SSL}{L}{SLL} & 0.187 & <0.001 & <0.001 \\ \hline
\tran{LSL}{S}{SLS} & <0.001 & <0.001 & <0.001 \\ \hline
\tran{LSL}{L}{SLL} & <0.001 & <0.001 & <0.001 \\ \hline
\tran{SLL}{S}{LLS} & 0.014 & <0.001 & <0.001 \\ \hline
\tran{SLL}{L}{LLL} & <0.001 & <0.001 & <0.001 \\ \hline
\tran{LLL}{S}{LLS} & <0.001 & <0.001 & <0.001 \\ \hline
\tran{LLL}{L}{LLL} & 0.627 & <0.001 & <0.001 \\ \hline
\end{tabular}

\label{table:norm_trans}
\end{table}

\begin{table}[!h]
\caption{\textbf{Complete word clusters from CODYMs of PCCRI corpus.} All words appearing in each of the six unsupervised clusters of 114 words, based on similarities in transition frequencies in a 2\textsuperscript{nd}-order CODYM. Words are parenthetically followed by the number of times each occurred in the PCCRI corpus. To comply with HIPPA privacy protection, all names and dates were anonymized; ``p\_name'' refers to the name of any patient or family member, ``u\_name'' refers to the name of an unknown speaker, and ``u\_date'` refers to any specific date. }
\begin{tabular}{l|l} 
 \hline
\textbf{Cluster Moniker (\# words)}& \textbf{Unique Words (\# occurrences in PCCRI corpus)}\\
\hline

\textbf{Strong Continuers (4 words)} &
\pbox{20cm}{\medskip hm (1987); hmm (1896); mm (1263); aha (277)\\}\\  \hline

\textbf{Moderate Continuers (10 words)}  &
\pbox{20cm}{\medskip yeah (15367); okay (12332); oh (3607); uh (3288); yes (2611); \\ huh (527); absolutely (402); wow (354); yup (334); yep (303)\\ }\\ \hline

\textbf{Weak Continuers (19 words)}  &
\pbox{20cm}{\medskip right (9663); no (6009); p\_name (2578); nice (1032); great (775); fine (679);\\ sorry (630); exactly (465);  alright (438);  meeting (266); true (247); \\ appreciate (210); correct (191); please (188); perfect (150); appetite (141);\\ beautiful (129); u\_date (118); boy (114)\\} \\ \hline

\textbf{Openers/Closers (7 words)} &
\pbox{20cm}{\medskip thank (1232); meet (565); hi (404); thanks (192);\\ welcome (135); hello (125); bye (111)\\} \\ \hline

\textbf{Clinical Talk (51 words):}  &
\pbox{20cm}{\medskip um (5993); also (1350); sort (978); may (907); comfort (685); \\ symptoms (623); treatments (525); whether (423); chemotherapy (409); \\ continue (337);  disease (267); methadone (263); focus (256); lungs (251);\\ means (233); treat (230); acting (228); system (194); symptom (186); \\machine (183);  oncology (182); likely (176); safe (170); fix (163); \\recommend (159);  show (158); illness (157); page (148); folks (145); \\ decided (139); controlled (135); outpatient (133); homes (130); setting (126); \\ clinic (125); further (124); belly (124); approach (124); death (122); \\ spot (120); consider (120); kept (117); ended (114); words (109);\\ example (108); icu (106); cpr (105); finally (105); depending (103); \\ depressed (102); stronger (101)\\} \\ \hline

\textbf{Potpourri (23 words):} &
\pbox{20cm}{\medskip well (5239); old (263); 5 (204); u\_name (178); hurts (176); tylenol (170);\\ totally (169); agree (165); friend (165); button (159); lives (144);\\ middle (142); concern (141); card (138); risk (137); cold (117); \\covered (116); fair (116); hit (115); confused (107); scary (104); \\mg (104); funny (103)\\}\\ \hline

\end{tabular}
\label{tab:full_word_clusters}
\end{table}

\begin{table}[!h]
\caption{\textbf{Mean transition frequencies in word clusters of PCCRI corpus.} Mean transition frequencies (\%) for  2\textsuperscript{nd}-order CODYMs for all words in the PCCRI corpus (ALL), and for the words in each of the six clusters: Strong Continuers (SC), Moderate Continuers (MC), Weak Continuers (WC), Openers/Closers (OC), Clinical Talk (CT), and PotPourri (PP).}
 \begin{tabular}{l|r|r|r|r|r|r|r} \hline
\textbf{Transition} & \textbf{ALL} & \textbf{SC} & \textbf{MC} & \textbf{WC} & \textbf{OC} & \textbf{CT} & \textbf{PP} \\ \hline
\tran{SS}{S}{SS} & 3.0 & 13.3 & 14.2 & 11.0 & 27.6 & 1.3 & 4.4 \\ \hline
\tran{LS}{S}{SS} & 2.7 & 7.4 & 8.5 & 7.0 & 12.7 & 1.3 & 4.0 \\ \hline
\tran{SL}{S}{LS} & 3.3 & 50.9 & 26.7 & 12.1 & 11.1 & 1.1 & 3.9 \\ \hline
\tran{LL}{S}{LS} & 2.0 & 17.6 & 13.8 & 6.3 & 4.4 & 0.9 & 2.9 \\ \hline
\tran{SS}{L}{SL} & 18.1 & 2.0 & 8.5 & 15.5 & 17.1 & 15.9 & 20.0 \\ \hline
\tran{LS}{L}{SL} & 35.8 & 2.0 & 11.1 & 21.4 & 12.5 & 47.7 & 24.5 \\ \hline
\tran{SL}{L}{LL} & 17.7 & 4.0 & 10.0 & 13.7 & 7.7 & 15.1 & 19.5 \\ \hline
\tran{LL}{L}{LL} & 17.4 & 2.9 & 7.2 & 12.9 & 6.9 & 16.8 & 20.8 \\ \hline
\end{tabular}

\label{table:cluster_freqs}
\end{table}
\clearpage

\end{document}